\pdfoutput=1

\documentclass[11pt]{article}

\usepackage[final]{acl}

\usepackage{times}
\usepackage{latexsym}

\usepackage[T1]{fontenc}

\usepackage[utf8]{inputenc}

\usepackage{microtype}

\usepackage{inconsolata}

\usepackage{graphicx}

\usepackage[utf8]{inputenc} 
\usepackage[T1]{fontenc}    
\usepackage{hyperref}       
\usepackage{url}            
\usepackage{booktabs}       
\usepackage{amsfonts}       
\usepackage{nicefrac}       
\usepackage{microtype}      
\usepackage{xcolor}         

\newcommand{\todot}[1]{\textcolor{black}{#1}}
\newcommand{\todop}[1]{\textcolor{black}{#1}}

\usepackage{tabularray}
\usepackage{color}
\usepackage{multirow}
\usepackage{amsmath}
\usepackage{makecell}
\usepackage{pifont}
\usepackage{fontawesome}
\definecolor{darkgreen}{RGB}{48,128,20}

\title{\textsc{GUIC}ourse: From General Vision Language Model to Versatile GUI Agent}

{\small
\author{
\\
\bf Wentong Chen$^1$\thanks{Equal Contribution.}, Junbo Cui$^2$$^{\ast}$, Jinyi Hu$^2$$^{\ast}$, Yujia Qin$^2$, Junjie Fang$^3$, \\
\bf Yue Zhao$^4$, Chongyi Wang$^5$, Jun Liu$^6$, Guirong Chen$^1$, Yupeng Huo$^1$, \\
\bf Yuan Yao$^{7,8}$\thanks{Corresponding Authors.} , Yankai Lin$^{1\dag}$, Zhiyuan Liu$^2$, Maosong Sun$^2$ \\
   \\
  $^1$Renmin University of China
  $^2$Tsinghua University
  $^3$Xiamen University\\
  $^4$BUPT 
  $^5$ModelBest 
  $^6$Chinese Academy of Sciences\\
  $^7$National University of Singapore
  $^8$Shanghai Qi Zhi Institute\\
  $^1$\texttt{cwt\_0139@ruc.edu.cn}
  $^2$\texttt{cuijb2000@gmail.com}
}
}

\begin{document}
\maketitle
\begin{abstract}
Utilizing Graphic User Interfaces (GUIs) for human-computer interaction is essential for accessing various digital tools. Recent advancements in Vision Language Models (VLMs) reveal significant potential for developing versatile agents that assist humans in navigating GUIs. However, current VLMs face challenges related to fundamental abilities, such as OCR and grounding, as well as a lack of knowledge about GUI elements functionalities and control methods. These limitations hinder their effectiveness as practical GUI agents.
To address these challenges, we introduce \textbf{GUICourse}, a series of datasets for training visual-based GUI agents using general VLMs. First, we enhance the OCR and grounding capabilities of VLMs using the \textbf{GUIEnv} dataset. Next, we enrich the GUI knowledge of VLMs using the \textbf{GUIAct} and \textbf{GUIChat} datasets. Our experiments demonstrate that even a small-sized GUI agent (with 3.1 billion parameters) performs effectively on both single-step and multi-step GUI tasks. We further fine-tune our GUI agents on other GUI tasks with different action spaces (AITW and Mind2Web), and the results show that our agents are better than their baseline VLMs. Additionally, we analyze the impact of OCR and grounding capabilities through an ablation study, revealing a positive correlation with GUI navigation ability. 
Our source codes and datasets are released at \url{https://github.com/RUCBM/GUICourse}.

\end{abstract}

\section{Introduction}
\label{sec:intro}



Graphical User Interfaces (GUIs) are a pivotal medium for facilitating human-computer interactions, playing a crucial role across diverse applications. GUI agents are designed to complete complex tasks in these GUI systems, which can liberate humans from tedious and repetitive operations. Different GUI systems (\todot{e.g.,} websites and smartphones) usually use icons and images to convey specific information, making them suitable to be processed by visual-based agents. On the other hand, obtaining screenshots of GUI systems is straightforward, whereas accessing structured text is challenging (such as the code behind the iOS system). In this work, we focus on training visual-based GUI agents from general VLMs.

\begin{figure*}[tb]
  \centering
  \includegraphics[width=1.0\linewidth]{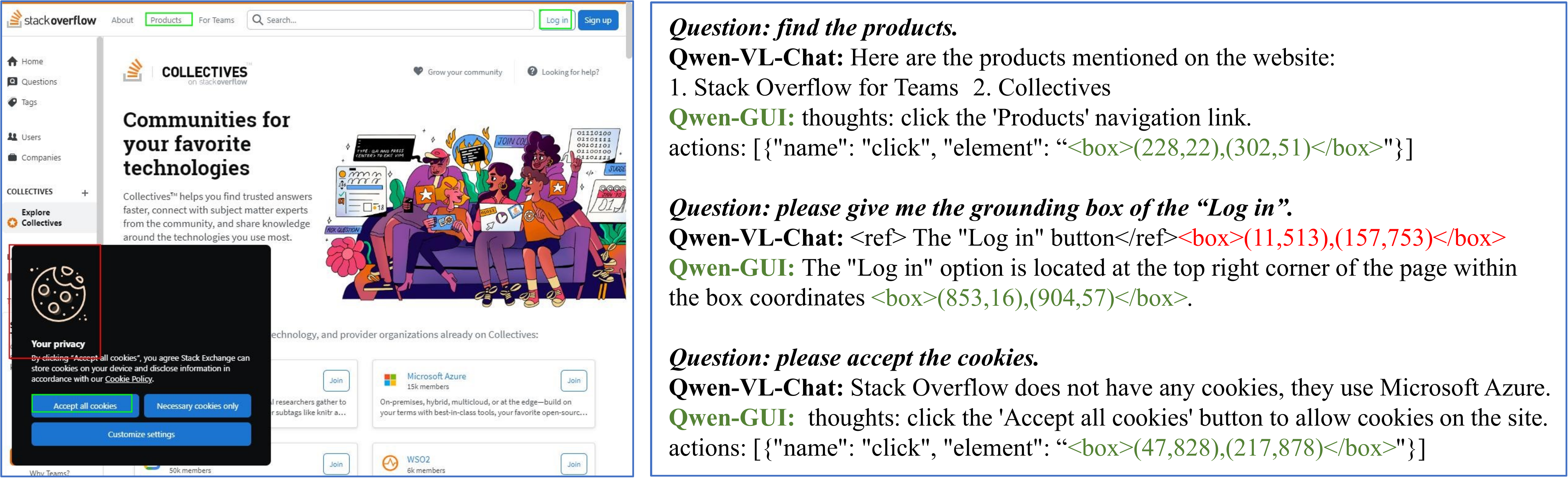}
  \caption{An example of several GUI tasks. Qwen-VL-Chat gives an incorrect bounding box for "Log in", which means its OCR and grounding capabilities are insufficient. Also, it can't finish GUI navigation tasks such as "find the products" and "accept the cookies" because it lacks the knowledge of  GUI control actions. Trained on our datasets, our GUI agent Qwen-GUI performs these tasks well.}
  \label{fig: qwen-example}
\end{figure*}

The performance of vision-based agents heavily relies on the fundamental capabilities and internal knowledge of their baseline VLMs. Although VLMs~\citep{liu2023visual, peng2023kosmos, zhu2023minigpt, li2023otterhd, hu2023large} demonstrate impressive capabilities in various tasks such as image captioning~\citep{pmlr-v202-li23q} and visual question answering~\citep{liu2023llava}, these capabilities may not suffice for developing efficient visual-based GUI agents. For instance, we investigate the Qwen-VL-Chat, a powerful open-sourced VLM, on its ability to finalize GUI navigation instructions. As illustrated in Figure~\ref{fig: qwen-example}, it shows the two challenges of current VLMs: {(1)} Their OCR and grounding abilities are not supported to locate the website elements or designated text accurately; {(2)} They lack a comprehensive understanding of the functions and control mechanisms of website widgets. To solve these problems, we provide a pipeline for training GUI agents based on general VLMs. This involves enhancing their OCR and grounding capabilities, followed by training them to perform GUI navigation tasks and engage in related conversations.

Our approach mainly employs data-driven methods for training diverse agents. To cultivate versatile GUI agents, we require a substantial amount of high-quality GUI navigation data enriched with visual information (e.g., screenshots). While previous works have contributed various datasets, they often suffer from one or more shortcomings: (1) the GUI environments may be overly simplistic and disconnected from real-world scenarios~\cite{liu2018reinforcement, humphreys2022data}; (2) the datasets tend to focus on narrow domains or scenarios~\cite{yao2022webshop, sun2022meta, rawles2023android}; and (3) the overall size of the datasets may be insufficient for effective training of GUI agents~\cite{deng2023mind2web}. In this work, we primarily introduce the \textbf{GUIAct} dataset, which addresses these challenges by providing 67k human-verified single-step and 5,696 human-annotated multi-step GUI navigation instructions across various websites. Additionally, we present the \textbf{GUIEnv} dataset, comprising 10 million website page-annotation pairs, and the \textbf{GUIChat} dataset, featuring 44k single-turn QA pairs and 6k multi-turn dialogues alongside website screenshots, to support our training pipeline.


\noindent{In summary, our contributions are as follows:}

(1) We establish a comprehensive pipeline and a series of datasets for building GUI agents based on VLMs. This pipeline consists of two learning stages aimed at enhancing fundamental abilities and acquiring knowledge specific to GUI systems.

(2) We train several visual-based GUI agents according to the pipeline and datasets. We find that even small-sized \todot{agents perform} effectively in GUI navigation tasks. Additionally, we demonstrate that integrating OCR and grounding data during the pre-training phase is helpful for these tasks.

\section{Related Works}
\label{sec: related works}

\subsection{Vision Language Models}

Currently, most Vision-Language Models (VLMs) use a bridge module to connect a vision encoder, such as CLIP-ViT~\citep{dosovitskiy2020image, radford2021learning}, and a language model, such as Vicuna~\citep{touvron2023llama} and Qwen~\citep{bai2023qwen}. The bridge module maps the visual features from the vision encoder into the embedding space of the large language model with an MLP layer, exemplified by LLaVA~\citep{liu2023llava}, or an attention-based resampler, such as BLIP-2~\citep{pmlr-v202-li23q} and Qwen-VL~\citep{Qwen-VL}. Uniquely, Fuyu-8B~\citep{fuyu-8b} removes the external vision encoder and uses a decoder-only transformer architecture with a unified space for textual tokens and pixel information, enabling Fuyu-8B to process images of any resolution. Based on this, subsequent research efforts try to enhance VLMs in various domains, including visual detail recognition~\citep{vstar, zhang2024exploring}, reliable responses~\citep{llava-rlhf, rlhf-v}, object grounding~\citep{peng2023kosmos}, multilingual capabilities~\citep{hu2023large} and model efficiency~\citep{zhu2024llava}. 
\todot{The current VLMs normally have OCR ability when recognizing sparse characters in small-size images~\citep{mishra2019ocr, singh2019towards} or dense text with similar fonts~\citep{blecher2023nougat}. However, they still have shortcomings when recognizing texts with different fonts in large screenshots and providing their accurate locations at the pixel level. In this work, we provide the GUIEnv dataset to improve the VLMs' OCR and grounding abilities with high-resolution website screenshots.}



\subsection{GUI Agents}

GUI agents are expected to help humans finish tasks on different GUI systems, such as simplified websites~\citep{shi2017world, liu2018reinforcement}, simulated environments~\citep{yao2022webshop, zhou2023webarena}, or real-world websites and smartphones~\citep{deng2023mind2web, rawles2023android}. Nowadays, the GUI agents are developing quickly due to the \todot{Large Models}, Big Data, and data-driven methods. Some agents~\citep{gao2023assistgpt, sun2023adaplanner, ma2023laser, kim2023language, xu2021grounding, zheng2023synapse, yang2023appagent} using the train-free method and \todot{depending on closed-sourced Large Models, such as GPT4 and GPT-4V~\citep{gpt4}}, to achieve GUI navigation with strategy design and prompt engineering. Also, some agents~\citep{humphreys2022data, nakano2021webgpt, qin2023webcpm} finish GUI tasks based on small-size open-sourced models, using data-driven methods (\todot{e.g., }SFT).

Perception and control are the two important points for GUI agents. According to the perception modalities of environments, the GUI agents can split into text-only~\citep{gur2023real, deng2023mind2web}, image-text combined~\citep{humphreys2022data, furuta2023multimodal}, and vision-only~\citep{shaw2023pixels, zhan2023you, hong2023cogagent} methods. As for the control, GUI agents can split into position-free and position-needed methods. The position-free methods~\cite{deng2023mind2web, yao2022webshop} generally provide indexes for every element in the GUI systems, and agents can finish actions by multiple choices. While the position-needed methods~\cite{shaw2023pixels, zhan2023you, hong2023cogagent} mean agents need to give the position information to finish some actions, such as "click" and "tap". \todop{Using text-based perception or position-free actions requires extracting element information from GUI systems. That might be difficult to obtain due to system permission restrictions. In contrast, vision-based information (e.g., screenshots) is readily available, and position-based actions align with the original design of GUI interactions. In this work, we aim to train GUI agents with vision-only inputs and position-based actions.}

 \section{GUICourse}
\label{sec: datasets}

In this section, we introduce GUICourse, a series of datasets for improving VLMs' OCR and grounding abilities, enhancing VLMs' GUI knowledge, and helping GUI agents' interaction. As shown in Table~\ref{tab: GUICourse}, our GUICourse includes GUIEnv, GUIAct, and GUIChat three partitions. Examples from the \textbf{GUICourse} datasets are provided in Appendix~\ref{sec: data-examples}. We employ a combination of LLM-based auto-annotation and human annotation to construct our datasets. More details of the annotation process (\todot{e.g., }the prompts and costs) are shown in Appendix~\ref{sec: prompts} and Appendix~\ref{sec: handbook}.


\begin{table}[tb]
  \small
  \centering
  \setlength\tabcolsep{1.5mm}{
  \begin{tabular}{lllr}
    \toprule
    Name & Subset & Source & Num of Ins. \\
    \midrule
    \multirow{2}{*}{GUIEnv} & global  & C4 & 10M \\
    & local & C4 & 0.7M \\
    \multirow{2}{*}{GUIAct} & web-single & GPT-4V/Human & 67k  \\
    & web-multi & Human & 5,696  \\
    & smartphone &AITW-general & 9,157  \\
    \multirow{2}{*}{GUIChat} & single-turn & GPT-4 & 44k  \\
    & multi-turn  & GPT-4 & 6k \\
    \bottomrule
  \end{tabular}}
  \caption{The overview of datasets we contributed for multi-stage GUI Agent training.}
  \label{tab: GUICourse}
\end{table}

\begin{table*}[tb]
  \small
  \centering
  \begin{tabular}{lrrrrcr}
    \toprule
     \textbf{Dataset} & \textbf{Dom.} & \textbf{Env.} & \textbf{Inst.} & \textbf{Avg. Turns} & \textbf{Inst. Level} & \textbf{Scenarios}\\
    \midrule
    MiniWoB++~\citep{liu2018reinforcement} & - & 100 & 100 & 3.6 & Low-level & Simplified Mob./Webs \\
    Rico~\citep{deka2017rico} & 27 & 9,772 & 10,811 & - & - & (Traces) Mob. Apps \\
    UIBert~\citep{bai2021uibert} & - & - & 16,660 & 1.0 & Low-level & Mob. Apps\\
    UGIF~\citep{venkatesh2022ugif} &  - & 12 & 523 & 5.3 & High \& low & Mob. Apps\\
    WebShop~\citep{yao2022webshop}& 1 & 1 & 12k & 11.3 & High-level & Shopping Webs \\
    RUSS~\citep{xu2021grounding} & - & 22 & 80 & - & High \& low &  Real-world Webs\\
    PixelHelp~\citep{li2020mapping} & 4 & 4 & 187 & - & High \& low&  Mobile Apps\\
    META-GUI~\citep{sun2022metagui} & 6 & 11 & 1,125 & 4.3 & High-level & (Dialogues) Mob. Apps \\
    MoTIF~\citep{burns2022dataset} & 15 & 125 & 756 & - &  High \& low & Mobile Apps\\
    MIND2WEB~\citep{deng2023mind2web} & 5/31 & 137 & 2,350  & 7.3 & High-level & Real-world Webs\\
    AITW~\citep{rawles2023android} & - & 357 & 715k & 6.5 & High \& low & Android/Apps \\
    WEBLINX~\citep{lù2024weblinx} & - & 155 & 2,337 & 43.0 & High-level & (Dialogues) Webs\\
    AndroidControl~\citep{li2024effects} & - & 833 & 15,283 & 4.8 & High \& low &  Mobile Apps\\
    GUI-World~\citep{chen2024gui} & - & - & 12,379 & - & High-level & Web./Mob./Desk./XR \\
    \midrule
    \textbf{GUIAct (web-single)} & 50 & 13k & 67k  & 1.0 & Low-level & Real-world Webs \\
    \textbf{GUIAct (web-multi)} & 8/32 & 121 & 5,696  & 7.9 & High-level & Real-world Webs \\
    \bottomrule
  \end{tabular}
  \caption{Statistics of the GUIAct dataset compared with existing datasets. The columns indicate: the number of app/website domains, environments, and instructions (Dom., Env., and Inst.), the average number of turns per instance (Avg. Turns), instruction level, and scenarios. Our main contributions are the two largest subsets in website scenarios with both single-step and multi-step instructions.}
  \label{tab: GUIAct}
\end{table*}

\subsection{GUIEnv: A Large Scale Dataset for OCR and Grounding Abilities}

OCR and grounding are two fundamental abilities for visual-based GUI agents' perception and control. 
\todop{To augment these abilities, we create a large-scale dataset called GUIEnv, which includes two subsets named GUIEnv-global and GUIEnv-local. GUIEnv-global has 10M samples for the pre-training, and each sample is a long text with all the describable content on the full page, including text, grounding information, and layout sequence. GUIEnv-local has 0.7M samples for the SFT, and each sample is a QA pair on a designated region as the "text2bbox" or "bbox2text" task. The samples of the two subsets are shown in Figure~\ref{fig: GUIEnv-example}.}

\paragraph{Data Collection.} We collected 4M URLs \todot{from the Cleaned Common Crawl Corpus~\citep{2019t5} which is referred as C4}. Then, we employed Playwright~\footnote{https://github.com/microsoft/playwright} for rendering, ultimately producing 10M annotated \todot{screenshots~\citep{lee2023pix2struct}}. These screenshots correspond to the 10M samples of the \emph{\textbf{GUIEnv-global}} data. As for the \emph{\textbf{GUIEnv-local}} data, we select about 50k website screenshots with their elements from C4. These screenshots are usually huge because they include the websites' complete content. We process the data in three steps: (1) crop them to smaller partitions less than \(1920 \times 1080 \) resolution\todot{, (2)} remove images with elements less than 10\todot{, (3) randomly sample} 10 elements with texts and positions in each screenshot as the data of "text2bbox" and "bbox2text" tasks. Finally, we acquired 0.7M instructions for the GUIEnv-local data. \todop{}

\subsection{GUIAct: A GUI Navigation Dataset with Single-step and Multi-step Instructions}

\begin{figure*}[tb]
  \centering
  \includegraphics[width=1.0\linewidth]{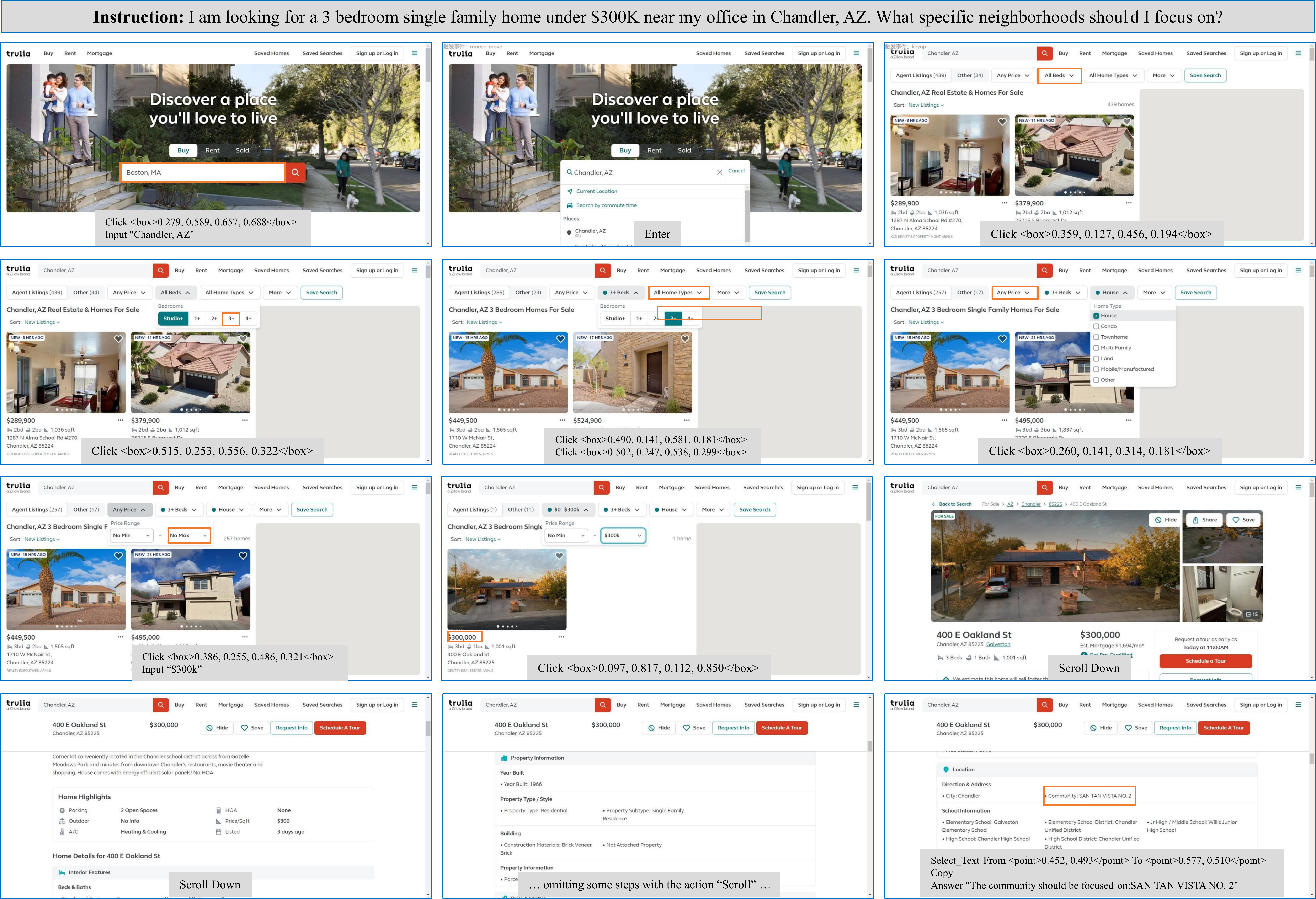}
  \caption{An example of our GUIAct dataset with 17 steps of action-screenshot pairs. Each step corresponds with one website screenshot and one or more actions. Some actions (e.g., “click”) have their location information, represented by a bounding box.}
  \label{fig: GUIAct}
\end{figure*}

We argue that GUI agents can learn GUI knowledge, such as the functions and the control methods of GUI elements from GUI navigation tasks. We provide the GUIAct dataset, which can be split into three partitions: \todop{single-step web navigation, multi-step web navigation, and smartphone navigation.} Our main contributions are two subsets in website scenarios: a single-step website navigation subset with 67k instructions and a multi-step website navigation subset with 5,696 instructions. As shown in Table~\ref{tab: GUIAct}, although there are many similar navigation tasks in mobile application scenarios, the two subsets are the largest navigation datasets in real-world website scenarios with both single-step and multi-step instructions. An example of our multi-step instruction is shown in Figure~\ref{fig: GUIAct}. The GUI agents only receive screenshots of GUI systems in the viewport, without any unseen elements, and output the action with position. We define a unified \textbf{action space} for different GUI systems, including 11 types of actions, and the details are shown in Appendix~\ref{sec: actions}. To enhance the scenario diversity of our dataset, we also convert a part of the samples in AITW depending on our action space and action format. \todop{We use the abbreviations "web-single", "web-multi", and "smartphone" to denote the three subsets in the subsequent context.}

\paragraph{Data Collection.} We contribute the \emph{\textbf{GUIAct (web-single)}} dataset in four steps: \emph{(1) Websites selection.} We use GPT-4 to gather 50 different scenarios (e.g., online shopping and E-learning) and hundreds of URLs. Then, we use these URLs as seeds to expand new websites by their hyperlinks, acquiring about 13k websites. \emph{(2) Captures acquisition.} We use web snapshot tools to respond to website HTML, interactive elements, and screenshots based on their URLs. \emph{(3) LLM-Auto Annotation.} We use GPT-4V to obtain single-step instruction-action pairs for each website. We give two images to GPT-4V in each request: an origin screenshot and a revised screenshot with interactive element identification. We get about 70k instruction-action pairs based on the 13k website screenshots. \emph{(4) Data checking by human.} We hire annotators to check the automatic instruction-action pairs' quality. The annotators are requested to revise the inaccurate items or abandon them if they are hard to revise. After that, the accuracy of GUIAct (web-single) data improves from 55\% to 92\%, based on our sampled results. Finally, we get around 67k single-step action instructions in website scenarios, corresponding to 67k training samples.

We contribute the \emph{\textbf{GUIAct (web-multi)}} dataset in three steps: \emph{(1) Websites selection.} We concept 8 top-level common web scenarios: Shopping, Diet, Stay, Travel, Education, Health, Job, and Entertainment. Based on these scenarios, we collect 32 sub-scenarios and 121 well-known websites. \emph{(2) Questions acquisition.} 
\todot{We collect 8k high-level instructions related to searching for specific information on the corresponding websites.} 
\emph{(3) Crowd-sourced Annotation.} We develop an annotation tool as a web browser plugin. Then, we hire annotators to execute operators to finish the instructions on the corresponding websites. If the instructions are inaccurate or have no answer, the annotators will revise or abandon them. Finally\todot{, we} get 5,696 multi-step action instructions in website scenarios, corresponding to 44k training samples.

We contribute the \emph{\textbf{GUIAct (smartphone)}} dataset using a subset of the AITW~\citep{rawles2023android} dataset. We select the data with the "General" tag and filter out the screenshots without the bottom navigation bar (we will convert the "go back" and "go home" actions to "tap"). Then, we convert the origin actions to our action space. Finally, we get 9,157 multi-step action instructions in smartphone scenarios, corresponding to 67k training samples.

\subsection{GUIChat: A Text-Rich Multi-Modal QA Dataset for Interaction}

Natural language interaction is important for better using GUI agents. It's necessary to mix conversation data during the training stage of GUI agents for better interaction. We introduce the GUIChat dataset, which includes lots of text-rich website screenshots and grounding boxes in dialogues.
This dataset has about 44k single-turn QA pairs and 6k multi-turn dialogues in four aspects: visual information queries, human-centric issues, world knowledge inquiries, and complex reasoning tasks. 

\paragraph{Data Collection.} We contribute this dataset in three steps: \emph{(1) Web Page Screenshot Acquisition.} Utilizing Playwright to render web pages and obtain screenshots, same as the collection process of GUIEnv. \emph{(2) Textual Representation Extraction.} Extracting essential structured information and text with coordinates from a redundant DOM tree. \emph{(3) Question-Answer Pair Generation.} Leveraging GPT-4 to construct question-answer pairs from the textual representations of the current website. The textual representations include the elements in the website with the position information. 

\section{Experiments}

We train several GUI agents depending on our training pipeline with our GUICourse (GUIEnv, GUIAct, and GUIChat datasets). First, we explore the performance of our GUI agents on our test datasets and common GUI tasks. Then, we do ablation studies of the whole training process based on MiniCPM-V, analyzing the influence of various factors. Finally, we use two examples to show the generalization of our GUI agents.

\begin{table*}[tb]
  \small
  \centering
  \setlength\tabcolsep{1mm}{
  \begin{tabular}{lcccccccccc}
    \toprule
    \multirow{2}{*}{\textbf{Agents}} & \multicolumn{3}{c}{Web-Single} & \multicolumn{3}{c}{Web-Multi} & \multicolumn{3}{c}{Smartphone} & Mean \\
    & Type EM & Cli.Acc & StepSR & Type EM & Cli.Acc & StepSR & Type EM & Cli.Acc & StepSR & StepSR\\
    \midrule
    \todot{GPT-4o-mini*} & 81.0 & 62.3 & 57.0 & 22.0 & 10.0 & 17.0 & 53.0 & 13.0 & 22.0 & 32.0 \\
    \midrule
    MiniCPM-GUI
    \scriptsize w/ GUIAct  & 89.1 & 52.4 & 55.2 & 62.3 & 42.0 & 48.6 & 64.5 & 29.6 & 43.3 & 49.0 \\
    \scriptsize + High Resolution  & 91.8 & 72.9 & 69.2 & 68.2 & 42.4 & 47.1 & 72.6 & 47.5 & 52.0 & 56.1 \\
    \scriptsize + GUIChat & \textbf{91.8} & \textbf{74.9} &\textbf{70.6} & 67.0 & 45.4 & \textbf{47.5} & 71.7 & 44.7 & 53.3 & 57.1 \\
    \midrule
    Fuyu-GUI & 90.1 & 67.1 & 63.5 & 68.8 & 51.2 & 47.1 & 72.1 & 29.1 & 40.4 & 50.4 \\
    Qwen-GUI & 90.9 & 69.4 & 66.7 & \textbf{68.9} & \textbf{52.5} & 46.8 & \textbf{73.0} & \textbf{55.7} & \textbf{58.1} & \textbf{57.2} \\
    \bottomrule
  \end{tabular}}
  \caption{The performance of our GUI agents on test datasets. \todot{GPT-4o-mini tested on 100 samples for each dataset.}}
  \label{tab: guiact-results}
\end{table*} 

\subsection{Training Settings}
Depending on our training pipeline, we first improve the VLMs' fundamental abilities (OCR and grounding ability) with the GUIEnv dataset. Then, we improve the VLMs' GUI knowledge and conversation ability using the GUIAct and GUIChat datasets. Considering the data quantity and format of GUICourse, we designate the GUIEnv-global data for pre-training, while the remaining datasets serve as supervised fine-tuning (SFT) data.

We train the MiniCPM-GUI agent based on \todot{MiniCPM-V~\citep{yao2024minicpm}} through the following steps: (1) We pre-train MiniCPM-V by integrating 10M samples from the GUIEnv-global dataset into the original pre-training data. (2) Subsequently, we continue the training in the SFT stage by merging approximately 70k samples from the GUIEnv-local dataset, along with all samples from the GUIAct and GUIChat datasets. We also train two versions of agents with different resolutions. \todop{The default resolution of MiniCPM-V is $448 \times 448$. We also have a high-resolution version with $1344 \times 1344$ by the flexible patching and slicing strategy~\citep{ye2023ureader}.}

Additionally, we train other agents, Qwen-GUI and Fuyu-GUI, based on Qwen-VL~\cite{Qwen-VL} and Fuyu-8B~\cite{fuyu-8b}, respectively. However, unlike the training of MiniCPM-GUI,  we only train them with the SFT data.

\subsection{Performance on GUI Navigation Tasks}
We evaluate our GUI agents through two methods: First, we test the performance of our agents on the test datasets split by ourselves. We split about 1.4k, 1k, and 2k samples from the "web-single", "web-multi", and "smartphone" partitions in GUIAct as test data. Then, we test the performance of our agents on other GUI navigation tasks with different action spaces, such as Mind2Web~\cite{deng2023mind2web} and AITW~\cite{rawles2023android}. We finetune our GUI agents and their baseline VLMs on the new GUI tasks. If our agents perform better, it demonstrates that they have learned more common GUI knowledge than baseline VLMs.

\paragraph{Results on our test datasets.} 
 We use the type exact match score (Type EM), click accuracy (Cli. Acc), and step success rate (StepSR) as metrics. Type EM is the accuracy of the actions' names. Cli.Acc is the average score of the success rate of the "click" and "tap" actions, which reflects the grounding ability. StepSR is the average score of all the actions' success rates. More details are shown in Appendix~\ref{sec: evaluation details}.
 
 As shown in Table~\ref{tab: guiact-results}, we can find that: (1) Small-size models can perform effectively on visual-based GUI navigation tasks. Qwen-GUI (9.6B) achieves the highest mean step success rate of $57.2$ across the three datasets, while MiniCPM-GUI (3.1B) delivers comparable results with a mean step success rate of $57.1$. (2) High resolution is important for GUI navigation tasks. 
 The mean step success rate improves to $56.1$ from $49.0$. (3) Our GUIChat dataset is helpful for tasks in website scenarios, although its main goal is to improve the interaction ability of GUI agents. The mean step success rate improved by \todot{$1$ point}.

\begin{table}[tb]
  \small
  \centering
  \setlength\tabcolsep{0.6mm}{
  \begin{tabular}{lccccccc}
    \toprule
    \multirow{2}{*}{\textbf{Agents}} & \multicolumn{2}{c}{Cross-Task} & \multicolumn{2}{c}{Cross-Website} & \multicolumn{2}{c}{Cross-Domain}\\
    & \scriptsize Ele.Acc & \scriptsize StepSR & \scriptsize Ele.Acc & \scriptsize StepSR & \scriptsize Ele.Acc & \scriptsize StepSR \\
    \midrule
    \scriptsize SeeClick & 28.3 & 25.5 & 21.4 & 16.4 & 23.2 & 20.8 \\
    \midrule
    \scriptsize Qwen-VL & 23.2 & 20.3 & 16.8 & 14.0 & 14.1& 12.3 \\
    \scriptsize Qwen-GUI & \textbf{27.9} & \textbf{24.4} & \textbf{19.3} & \textbf{15.6} & \textbf{20.5} & \textbf{17.5}\\
    \midrule
    \scriptsize Fuyu-8B & 8.3 & \ \ 6.6 & \ \ 4.8  & \ \ 4.0 & \ \ 3.6& \ \ 3.0\\
    \scriptsize Fuyu-GUI & \textbf{19.1} & \textbf{15.6} & \textbf{13.9} & \textbf{12.2} & \textbf{14.2} & \textbf{11.7} \\
    \midrule
    \scriptsize MiniCPM-V & 11.0  & \ \ 8.5 & \ \ 8.2 & \ \ 6.0 & \ \ 6.5 & \ \ 5.2\\
    \scriptsize MiniCPM-GUI & \textbf{23.8} & \textbf{20.8} & \textbf{20.3} & \textbf{17.3} & \textbf{17.9} & \textbf{14.6} \\
    \bottomrule
  \end{tabular}}
  \caption{Results of our GUI agents on Mind2Web.}
  \label{tab: mind2web-results}
\end{table} 

\begin{table}[tb]
  \small
  \centering
  \setlength\tabcolsep{0.6mm}{
  \begin{tabular}{lrccccc}
    \toprule
    \textbf{Models} & Install & Goo.Apps & Single & WebShop. & Overall \\
    \midrule
    \scriptsize SeeClick  & 66.4 & 54.9 & 63.5 & 57.6 & 60.6 \\
    \midrule
    \scriptsize Qwen-VL   & \textbf{70.4} & 57.8 & 70.1 & 64.7 & 65.8  \\
    \scriptsize Qwen-GUI  & 70.3 & \textbf{61.2} & \textbf{71.6} & \textbf{66.1} & \textbf{67.3} \\
    \midrule
    \scriptsize Fuyu-8B & 45.9 & 40.0 & \textbf{47.2} & 40.8 & 43.5 \\
    \scriptsize Fuyu-GUI  & \textbf{50.9} & \textbf{41.6} & 45.7 & \textbf{43.8} & \textbf{45.5}  \\
    \midrule
    \scriptsize MiniCPM-V & 50.2 & 45.1 & 56.2 & 44.0 & 48.9 \\
    \scriptsize MiniCPM-GUI  & \textbf{62.3} & \textbf{46.5} & \textbf{67.3} & \textbf{57.5} & \textbf{58.4} \\
    \bottomrule
  \end{tabular}}
  \caption{Results of our GUI agents on the AITW.}
  \label{tab: aitw-results}
\end{table}

\begin{table*}[tb]
  \small
  \centering
  \setlength\tabcolsep{1.6mm}{
  \begin{tabular}{lcccccccc}
    \toprule
    \multirow{2}{*}{\textbf{Amount}} & \multicolumn{2}{c}{Bbox2Text} & \multicolumn{3}{c}{Text2Bbox} & \multicolumn{3}{c}{Web-Single} \\
    & EM Score & F1 Score & IoU@0.2 & IoU@0.5 & IoU@0.7 & Type EM & Cli.Acc & StepSR \\
    \midrule
    w/o GUIEnv & 14.92 & 30.49 & \ \ 9.87 & \ \ 2.15 & \ \ 0.32 & 92.70 & 49.80 & 52.84 \\
    w/ 2.5M GUIEnv & 35.87 & 57.38 & 51.39 & 25.32 & \ \ 9.76 & 91.84 & 66.40 & 64.82 \\
    w/ 5.0M GUIEnv & 40.21 & 62.61 & 60.73 & 35.09 & 16.09 & \textbf{92.98} & 71.20 & 68.30 \\ 
    w/ 7.5M GUIEnv & 43.60 & 63.46 & 65.34 & 40.67 & 20.06 & 91.56 & 71.50 & 68.58 \\
    w/ \ 10M GUIEnv & \textbf{44.12} & \textbf{64.78} & \textbf{68.02} & \textbf{47.96} & \textbf{23.28} & 91.77 & \textbf{74.90} & \textbf{70.57} \\
    \bottomrule
  \end{tabular}}
  \caption{The performance of MiniCPM-GUI with different amounts of the GUIEnv data.}
  \label{tab: guienv-results}
\end{table*}

\paragraph{Results on \todot{other} GUI navigation datasets.}
Mind2Web and AITW are two commonly used GUI tasks. Mind2Web has 2000 tasks with high-level instructions in website scenarios. AITW has 30k instructions and corresponding 715k operation trajectories in smartphone scenarios. To keep consistency with previous works, we choose the processed version of these two tasks by SeeClick~\citep{cheng2024seeclick} and use the same evaluation methods as it does. We remove the "General" column of AITW in the evaluation because we convert and use a subset of the AITW-General dataset into our GUIAct dataset.

We mainly compared our GUI agents with their baseline VLMs. \todot{The main metric is step success rate (StepSR) and we also use element accuracy (Ele.Acc) for Mind2Web, a metric to evaluate whether the action's position parameters are correct.}. The results in Table~\ref{tab: mind2web-results} and Table~\ref{tab: aitw-results} demonstrate that GUICourse helps GUI agents achieve better performance than their baseline VLMs. (1)  In the Mind2Web task, Qwen-GUI exhibits a remarkable increase of $2$-$5$ points in step success rate compared to its baseline Qwen-VL. Fuyu-8B and MiniCPM-V, which have weaker fundamental abilities, show more improvements of approximately $10$ points. (2) In the AITW task, our agents still have better performance than their baseline VLMs. \todot{However,} these improvements are relatively smaller compared to the Mind2Web task, likely because we use more data (\todot{e.g., }GUIEnv and GUIChat) in the website scenario.

\subsection{Ablation Study}

In this section, we show the influence of VLMs' OCR and grounding abilities on simple GUI navigation tasks. We adjust the amount of the GUIEnv-global dataset in the pertaining data and analyze the final performances of MiniCPM-GUI.

\paragraph{Metrics of OCR and grounding abilities.} We randomly sample 1.8k samples from the GUIEnv-local dataset as the test data of \emph{\textbf{Bbox2Text and Text2Bbox}} tasks, testing the OCR and grounding ability of agents. We evaluate the Bbox2Text task by EM and F1 scores, which is the same as the SQuAD MRC~\citep{rajpurkar2016squad} task. We evaluate the Text2Bbox task by IoU scores (IoU@0.2, 0.5, and 0.7).

\paragraph{Results.} The performances of MiniCPM-GUI with different settings are shown in Table~\ref{tab: guienv-results}. We demonstrate that our GUIEnv dataset is useful for \todop{improving the OCR and grounding ability} of VLMs and the GUI navigation ability. As shown in Table~\ref{tab: guienv-results}, the scores of the "text2bbox" and "bbox2text" tasks, as well as the performance on GUI navigation tasks (web-single) is rising during the increase in the amount of GUIEnv data. We can see that the improvement is huge from "without GUIEnv" to "2.5M GUIEnv", which demonstrates that OCR and grounding abilities are necessary. The amount of GUIEnv data does not influence the "Type EM" because predicting action names is position-free.

\subsection{Case Study}

\begin{figure*}[!t]
  \centering
  \includegraphics[width=1.0\linewidth]{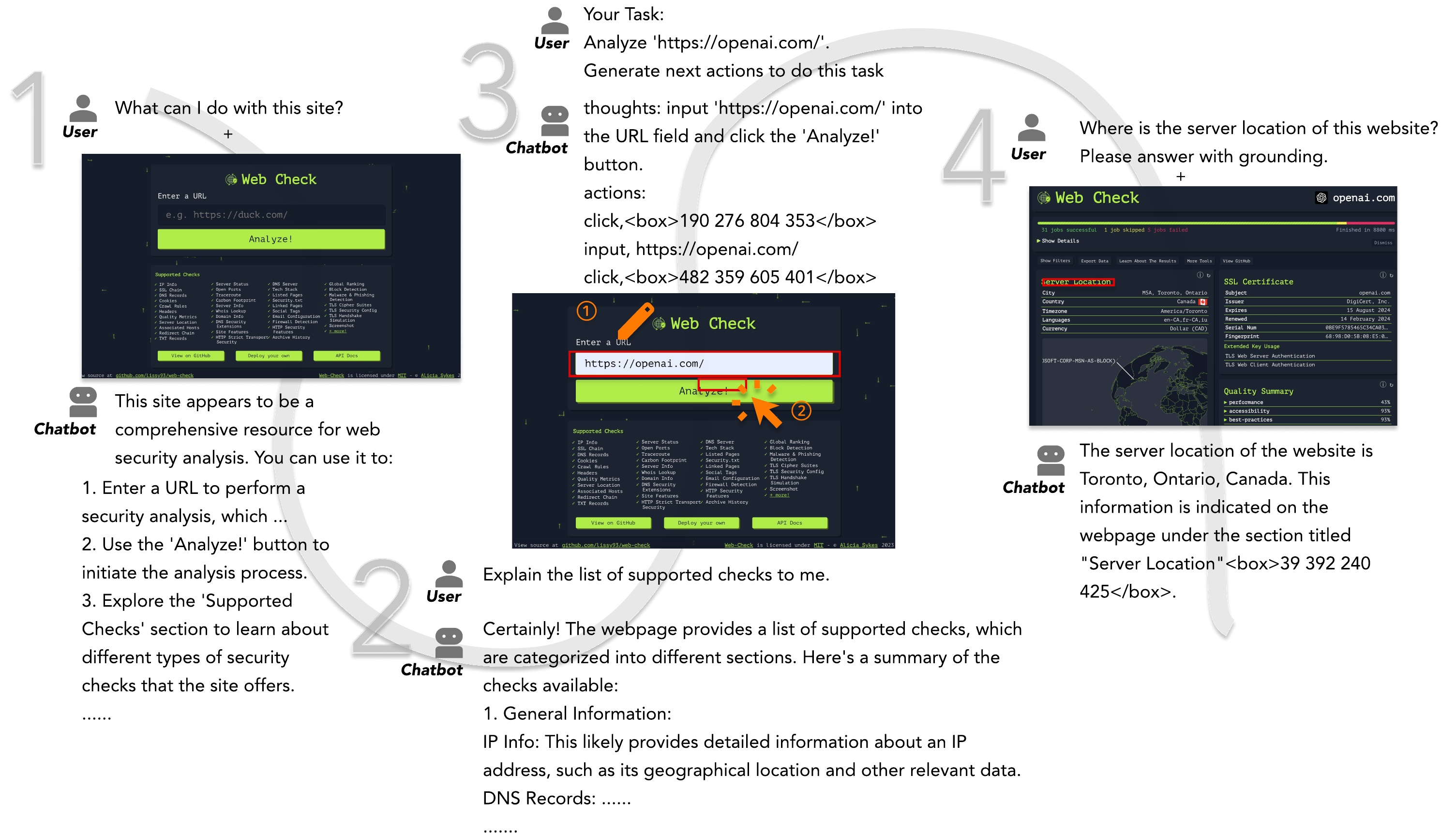}
  \caption{An example of a multi-turn conversation with our GUI agent, MiniCPM-GUI. We first ask the agent about possible operations on this website, then let it explain the terminologies in the screenshot. After that, we test the grounding ability of the agent by giving it a GUI navigation instruction and an information search question. 
  }
  \label{fig: chat with GUI Agent}
\end{figure*}

\paragraph{Multi-Turn Dialogue.} As shown in Figure~\ref{fig: chat with GUI Agent}, we show a multi-turn dialogue example of our GUI agents. This example uses several unseen website screenshots and illustrates its proficiency across four key areas: (1) Analyzing the functions of GUI systems' elements; (2) Answering questions using its internal knowledge; (3) Creating actionable steps to accomplish specific tasks; (4) Extracting textual information from the screenshot along with its coordinates through grounding.

\begin{figure*}[!t]
  \centering
  \includegraphics[width=1.0\linewidth]{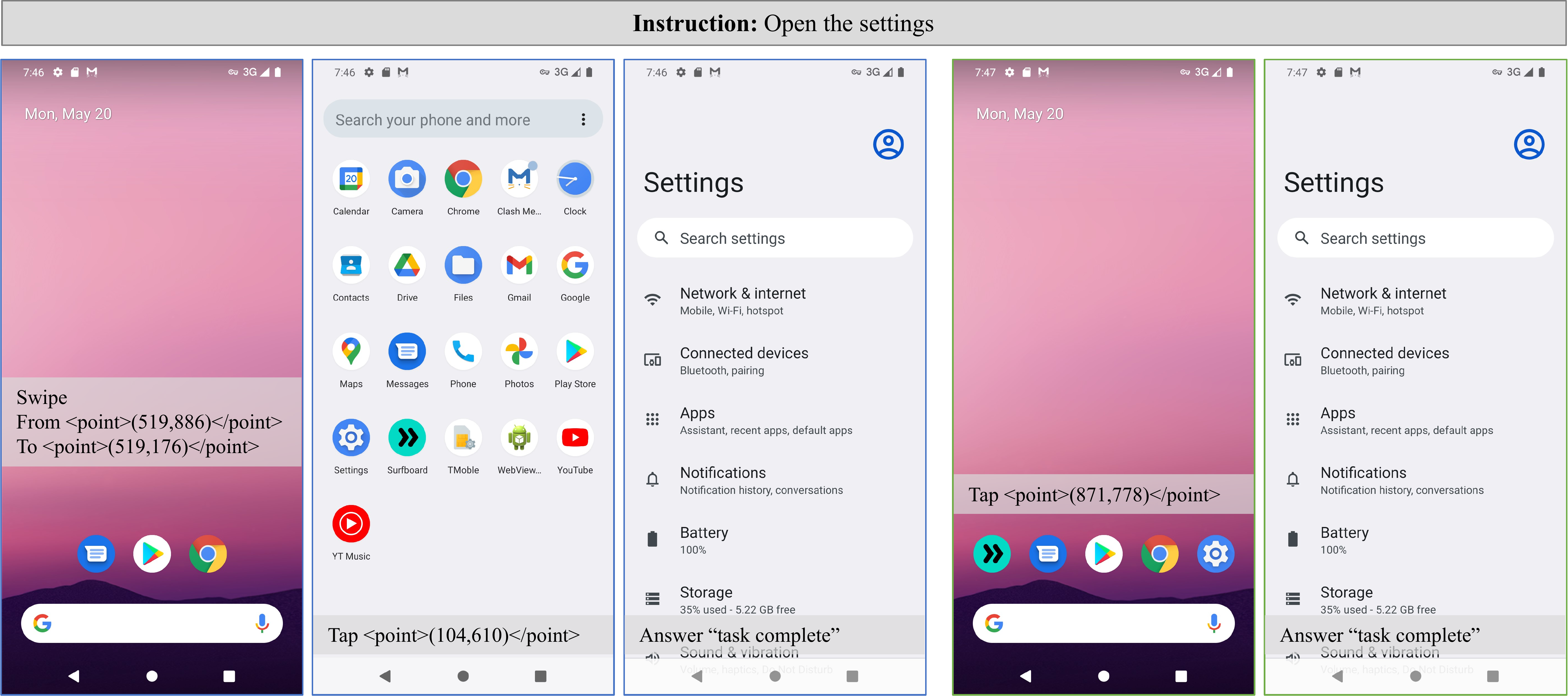}
  \caption{An example of our GUI agent executing the same instruction with different environments. The numbers in the "<point></point>" are the related positions from 0 to 1000.}
  \label{fig: robustness-testing}
\end{figure*}

\paragraph{Simulated Environment.}
To evaluate the robustness of our GUI agents, we use a simulated smartphone environment by Android Studio\footnote{https://developer.android.com/} to do interactive testing. As shown in Figure~\ref{fig: robustness-testing}, we let our GUI agent execute the same instruction with different initial environments. The execution processes are shown in the the left three and the right two screenshots respectively. When there is no "setting" icon on the initial environment, our GUI agent swipes first to find the "setting" icon. While the "setting" icon appears on the initial environment, our GUI agent taps directly. That means our agents can distinguish the detailed changes and know the icon of "settings", not only learn the fixed action sequences in the training data. A long-chain example is shown in Appendix~\ref{sec: android}.

\subsection{Error analysis}

\todop{In this section, we analyze the error cases of our MiniCPM-GUI on one of our testing tasks (web-multi). We selected 50 error samples and manually reviewed them. After analysis, we categorized the errors into two main types: \textbf{\textit{(1) False Errors.}} The model’s output is reasonable but does not match the predefined golden answer. \textbf{\textit{(2) True Errors.}} The model’s output is unreasonable, including \textit{Action Type Errors} (such as "click" and "swipe") and \textit{Action Position Errors} (the grounding boxes). }

\todop{Out of the 50 error samples analyzed, 13 samples were classified as false errors, indicating that the behavior was reasonable. There are 22 error samples related to action type errors, and 15 error samples are action position errors. We have identified some key insights from the error analysis:}

\todop{(1) For the same instruction and initial screenshot, there can be multiple valid approaches to achieve the task. Static evaluation has a drawback in that some reasonable actions may be incorrectly judged as errors, referred to as false errors.}

\todop{(2) Our GUI agents struggle to complete instructions when the screenshot content is complex, which can’t give a correct action, referred to as true errors.}

\section{Conclusion}
In this work, we provide a comprehensive pipeline to train visual-based GUI agents from general VLMs. We construct GUICourse, a group of complete datasets (GUIEnv, GUIAct, and GUIChat) to help train GUI agents from general VLMs from three aspects respectively: improving the OCR and grounding abilities of VLMs, enhancing VLMs' GUI knowledge, and improving GUI agents' interaction ability. We train several GUI agents, evaluate them on GUI navigation tasks, and try them on simulated Android environments. Meanwhile, we conduct ablation studies to show the influence of OCR and grounding abilities. 

\section*{Limitations} We use pretraining and supervised fine-tuning for training GUI agents using various VLMs. However, these might not be enough to achieve assistant-level GUI agents. We are considering the reinforcement learning methods such as RLHF in the future. We provide training data about website environments and fuse data about smartphone environments in our training. However, to build a general GUIAgent, only website and smartphone scenarios are not enough. We would like to build or gather more GUI instruction data about computer systems and professional software in the future.

\section*{Acknowledgements}
We sincerely thank all the anonymous reviewers and (S)ACs for their constructive comments and helpful suggestions. This work was supported by The National Natural Science Foundation of China (No.\ 62376273 and No.U2436209) and Shanghai Qi Zhi Institute Innovation Program SQZ202410.

\bibliography{custom}

\begin{thebibliography}{59}
\providecommand{\natexlab}[1]{#1}

\bibitem[{Bai et~al.(2021)Bai, Zang, Xu, Sunkara, Rastogi, Chen et~al.}]{bai2021uibert}
Chongyang Bai, Xiaoxue Zang, Ying Xu, Srinivas Sunkara, Abhinav Rastogi, Jindong Chen, et~al. 2021.
\newblock Uibert: Learning generic multimodal representations for ui understanding.
\newblock \emph{arXiv preprint arXiv:2107.13731}.

\bibitem[{Bai et~al.(2023{\natexlab{a}})Bai, Bai, Chu, Cui, Dang, Deng, Fan, Ge, Han, Huang et~al.}]{bai2023qwen}
Jinze Bai, Shuai Bai, Yunfei Chu, Zeyu Cui, Kai Dang, Xiaodong Deng, Yang Fan, Wenbin Ge, Yu~Han, Fei Huang, et~al. 2023{\natexlab{a}}.
\newblock Qwen technical report.
\newblock \emph{arXiv preprint arXiv:2309.16609}.

\bibitem[{Bai et~al.(2023{\natexlab{b}})Bai, Bai, Yang, Wang, Tan, Wang, Lin, Zhou, and Zhou}]{Qwen-VL}
Jinze Bai, Shuai Bai, Shusheng Yang, Shijie Wang, Sinan Tan, Peng Wang, Junyang Lin, Chang Zhou, and Jingren Zhou. 2023{\natexlab{b}}.
\newblock Qwen-vl: A frontier large vision-language model with versatile abilities.
\newblock \emph{arXiv preprint arXiv:2308.12966}.

\bibitem[{Bavishi et~al.(2023)Bavishi, Elsen, Hawthorne, Nye, Odena, Somani, and Ta\c{s}\i{}rlar}]{fuyu-8b}
Rohan Bavishi, Erich Elsen, Curtis Hawthorne, Maxwell Nye, Augustus Odena, Arushi Somani, and Sa\u{g}nak Ta\c{s}\i{}rlar. 2023.
\newblock \href {https://www.adept.ai/blog/fuyu-8b} {Introducing our multimodal models}.

\bibitem[{Blecher et~al.(2023)Blecher, Cucurull, Scialom, and Stojnic}]{blecher2023nougat}
Lukas Blecher, Guillem Cucurull, Thomas Scialom, and Robert Stojnic. 2023.
\newblock Nougat: Neural optical understanding for academic documents.
\newblock \emph{arXiv preprint arXiv:2308.13418}.

\bibitem[{Burns et~al.(2022)Burns, Arsan, Agrawal, Kumar, Saenko, and Plummer}]{burns2022dataset}
Andrea Burns, Deniz Arsan, Sanjna Agrawal, Ranjitha Kumar, Kate Saenko, and Bryan~A Plummer. 2022.
\newblock A dataset for interactive vision-language navigation with unknown command feasibility.
\newblock In \emph{European Conference on Computer Vision}, pages 312--328. Springer.

\bibitem[{Chen et~al.(2024)Chen, Huang, Wu, Tang, Chen, Bai, He, Wang, Zhou, Li et~al.}]{chen2024gui}
Dongping Chen, Yue Huang, Siyuan Wu, Jingyu Tang, Liuyi Chen, Yilin Bai, Zhigang He, Chenlong Wang, Huichi Zhou, Yiqiang Li, et~al. 2024.
\newblock Gui-world: A dataset for gui-oriented multimodal llm-based agents.
\newblock \emph{arXiv preprint arXiv:2406.10819}.

\bibitem[{Cheng et~al.(2024)Cheng, Sun, Chu, Xu, Li, Zhang, and Wu}]{cheng2024seeclick}
Kanzhi Cheng, Qiushi Sun, Yougang Chu, Fangzhi Xu, Yantao Li, Jianbing Zhang, and Zhiyong Wu. 2024.
\newblock Seeclick: Harnessing gui grounding for advanced visual gui agents.
\newblock \emph{arXiv preprint arXiv:2401.10935}.

\bibitem[{Deka et~al.(2017)Deka, Huang, Franzen, Hibschman, Afergan, Li, Nichols, and Kumar}]{deka2017rico}
Biplab Deka, Zifeng Huang, Chad Franzen, Joshua Hibschman, Daniel Afergan, Yang Li, Jeffrey Nichols, and Ranjitha Kumar. 2017.
\newblock Rico: A mobile app dataset for building data-driven design applications.
\newblock In \emph{Proceedings of the 30th annual ACM symposium on user interface software and technology}, pages 845--854.

\bibitem[{Deng et~al.(2023)Deng, Gu, Zheng, Chen, Stevens, Wang, Sun, and Su}]{deng2023mind2web}
Xiang Deng, Yu~Gu, Boyuan Zheng, Shijie Chen, Samuel Stevens, Boshi Wang, Huan Sun, and Yu~Su. 2023.
\newblock Mind2web: Towards a generalist agent for the web.
\newblock \emph{arXiv preprint arXiv:2306.06070}.

\bibitem[{Dosovitskiy et~al.(2020)Dosovitskiy, Beyer, Kolesnikov, Weissenborn, Zhai, Unterthiner, Dehghani, Minderer, Heigold, Gelly et~al.}]{dosovitskiy2020image}
Alexey Dosovitskiy, Lucas Beyer, Alexander Kolesnikov, Dirk Weissenborn, Xiaohua Zhai, Thomas Unterthiner, Mostafa Dehghani, Matthias Minderer, Georg Heigold, Sylvain Gelly, et~al. 2020.
\newblock An image is worth 16x16 words: Transformers for image recognition at scale.
\newblock \emph{arXiv preprint arXiv:2010.11929}.

\bibitem[{Furuta et~al.(2023)Furuta, Nachum, Lee, Matsuo, Gu, and Gur}]{furuta2023multimodal}
Hiroki Furuta, Ofir Nachum, Kuang-Huei Lee, Yutaka Matsuo, Shixiang~Shane Gu, and Izzeddin Gur. 2023.
\newblock Multimodal web navigation with instruction-finetuned foundation models.
\newblock \emph{arXiv preprint arXiv:2305.11854}.

\bibitem[{Gao et~al.(2023)Gao, Ji, Zhou, Lin, Chen, Fan, and Shou}]{gao2023assistgpt}
Difei Gao, Lei Ji, Luowei Zhou, Kevin~Qinghong Lin, Joya Chen, Zihan Fan, and Mike~Zheng Shou. 2023.
\newblock Assistgpt: A general multi-modal assistant that can plan, execute, inspect, and learn.
\newblock \emph{arXiv preprint arXiv:2306.08640}.

\bibitem[{Gur et~al.(2023)Gur, Furuta, Huang, Safdari, Matsuo, Eck, and Faust}]{gur2023real}
Izzeddin Gur, Hiroki Furuta, Austin Huang, Mustafa Safdari, Yutaka Matsuo, Douglas Eck, and Aleksandra Faust. 2023.
\newblock A real-world webagent with planning, long context understanding, and program synthesis.
\newblock \emph{arXiv preprint arXiv:2307.12856}.

\bibitem[{Hong et~al.(2023)Hong, Wang, Lv, Xu, Yu, Ji, Wang, Wang, Dong, Ding et~al.}]{hong2023cogagent}
Wenyi Hong, Weihan Wang, Qingsong Lv, Jiazheng Xu, Wenmeng Yu, Junhui Ji, Yan Wang, Zihan Wang, Yuxiao Dong, Ming Ding, et~al. 2023.
\newblock Cogagent: A visual language model for gui agents.
\newblock \emph{arXiv preprint arXiv:2312.08914}.

\bibitem[{Hu et~al.(2023)Hu, Yao, Wang, Wang, Pan, Chen, Yu, Wu, Zhao, Zhang et~al.}]{hu2023large}
Jinyi Hu, Yuan Yao, Chongyi Wang, Shan Wang, Yinxu Pan, Qianyu Chen, Tianyu Yu, Hanghao Wu, Yue Zhao, Haoye Zhang, et~al. 2023.
\newblock Large multilingual models pivot zero-shot multimodal learning across languages.
\newblock \emph{arXiv preprint arXiv:2308.12038}.

\bibitem[{Humphreys et~al.(2022)Humphreys, Raposo, Pohlen, Thornton, Chhaparia, Muldal, Abramson, Georgiev, Santoro, and Lillicrap}]{humphreys2022data}
Peter~C Humphreys, David Raposo, Tobias Pohlen, Gregory Thornton, Rachita Chhaparia, Alistair Muldal, Josh Abramson, Petko Georgiev, Adam Santoro, and Timothy Lillicrap. 2022.
\newblock A data-driven approach for learning to control computers.
\newblock In \emph{International Conference on Machine Learning}, pages 9466--9482. PMLR.

\bibitem[{Kim et~al.(2023)Kim, Baldi, and McAleer}]{kim2023language}
Geunwoo Kim, Pierre Baldi, and Stephen McAleer. 2023.
\newblock Language models can solve computer tasks.
\newblock \emph{arXiv preprint arXiv:2303.17491}.

\bibitem[{Lee et~al.(2023)Lee, Joshi, Turc, Hu, Liu, Eisenschlos, Khandelwal, Shaw, Chang, and Toutanova}]{lee2023pix2struct}
Kenton Lee, Mandar Joshi, Iulia Turc, Hexiang Hu, Fangyu Liu, Julian Eisenschlos, Urvashi Khandelwal, Peter Shaw, Ming-Wei Chang, and Kristina Toutanova. 2023.
\newblock \href {https://arxiv.org/abs/2210.03347} {Pix2struct: Screenshot parsing as pretraining for visual language understanding}.
\newblock \emph{Preprint}, arXiv:2210.03347.

\bibitem[{Li et~al.(2023{\natexlab{a}})Li, Zhang, Yang, Zhang, Pu, and Liu}]{li2023otterhd}
Bo~Li, Peiyuan Zhang, Jingkang Yang, Yuanhan Zhang, Fanyi Pu, and Ziwei Liu. 2023{\natexlab{a}}.
\newblock Otterhd: A high-resolution multi-modality model.
\newblock \emph{arXiv preprint arXiv:2311.04219}.

\bibitem[{Li et~al.(2023{\natexlab{b}})Li, Li, Savarese, and Hoi}]{pmlr-v202-li23q}
Junnan Li, Dongxu Li, Silvio Savarese, and Steven Hoi. 2023{\natexlab{b}}.
\newblock \href {https://proceedings.mlr.press/v202/li23q.html} {{BLIP}-2: Bootstrapping language-image pre-training with frozen image encoders and large language models}.
\newblock In \emph{Proceedings of the 40th International Conference on Machine Learning}, volume 202 of \emph{Proceedings of Machine Learning Research}, pages 19730--19742. PMLR.

\bibitem[{Li et~al.(2024)Li, Bishop, Li, Rawles, Campbell-Ajala, Tyamagundlu, and Riva}]{li2024effects}
Wei Li, William Bishop, Alice Li, Chris Rawles, Folawiyo Campbell-Ajala, Divya Tyamagundlu, and Oriana Riva. 2024.
\newblock On the effects of data scale on computer control agents.
\newblock \emph{arXiv preprint arXiv:2406.03679}.

\bibitem[{Li et~al.(2020)Li, He, Zhou, Zhang, and Baldridge}]{li2020mapping}
Yang Li, Jiacong He, Xin Zhou, Yuan Zhang, and Jason Baldridge. 2020.
\newblock Mapping natural language instructions to mobile ui action sequences.
\newblock \emph{arXiv preprint arXiv:2005.03776}.

\bibitem[{Liu et~al.(2018)Liu, Guu, Pasupat, Shi, and Liang}]{liu2018reinforcement}
Evan~Zheran Liu, Kelvin Guu, Panupong Pasupat, Tianlin Shi, and Percy Liang. 2018.
\newblock Reinforcement learning on web interfaces using workflow-guided exploration.
\newblock \emph{arXiv preprint arXiv:1802.08802}.

\bibitem[{Liu et~al.(2023{\natexlab{a}})Liu, Li, Wu, and Lee}]{liu2023visual}
Haotian Liu, Chunyuan Li, Qingyang Wu, and Yong~Jae Lee. 2023{\natexlab{a}}.
\newblock Visual instruction tuning.
\newblock \emph{arXiv preprint arXiv:2304.08485}.

\bibitem[{Liu et~al.(2023{\natexlab{b}})Liu, Cheng, Liu, Zhang, Li, Ren, Zou, Yang, Su, Zhu et~al.}]{liu2023llava}
Shilong Liu, Hao Cheng, Haotian Liu, Hao Zhang, Feng Li, Tianhe Ren, Xueyan Zou, Jianwei Yang, Hang Su, Jun Zhu, et~al. 2023{\natexlab{b}}.
\newblock Llava-plus: Learning to use tools for creating multimodal agents.
\newblock \emph{arXiv preprint arXiv:2311.05437}.

\bibitem[{Lù et~al.(2024)Lù, Kasner, and Reddy}]{lù2024weblinx}
Xing~Han Lù, Zdeněk Kasner, and Siva Reddy. 2024.
\newblock \href {https://arxiv.org/abs/2402.05930} {Weblinx: Real-world website navigation with multi-turn dialogue}.
\newblock \emph{Preprint}, arXiv:2402.05930.

\bibitem[{Ma et~al.(2023)Ma, Zhang, Wang, Pan, and Yu}]{ma2023laser}
Kaixin Ma, Hongming Zhang, Hongwei Wang, Xiaoman Pan, and Dong Yu. 2023.
\newblock Laser: Llm agent with state-space exploration for web navigation.
\newblock \emph{arXiv preprint arXiv:2309.08172}.

\bibitem[{Mishra et~al.(2019)Mishra, Shekhar, Singh, and Chakraborty}]{mishra2019ocr}
Anand Mishra, Shashank Shekhar, Ajeet~Kumar Singh, and Anirban Chakraborty. 2019.
\newblock Ocr-vqa: Visual question answering by reading text in images.
\newblock In \emph{2019 international conference on document analysis and recognition (ICDAR)}, pages 947--952. IEEE.

\bibitem[{Nakano et~al.(2021)Nakano, Hilton, Balaji, Wu, Ouyang, Kim, Hesse, Jain, Kosaraju, Saunders et~al.}]{nakano2021webgpt}
Reiichiro Nakano, Jacob Hilton, Suchir Balaji, Jeff Wu, Long Ouyang, Christina Kim, Christopher Hesse, Shantanu Jain, Vineet Kosaraju, William Saunders, et~al. 2021.
\newblock Webgpt: Browser-assisted question-answering with human feedback.
\newblock \emph{arXiv preprint arXiv:2112.09332}.

\bibitem[{OpenAI(2023)}]{gpt4}
OpenAI. 2023.
\newblock Gpt-4 technical report.

\bibitem[{Peng et~al.(2023)Peng, Wang, Dong, Hao, Huang, Ma, and Wei}]{peng2023kosmos}
Zhiliang Peng, Wenhui Wang, Li~Dong, Yaru Hao, Shaohan Huang, Shuming Ma, and Furu Wei. 2023.
\newblock Kosmos-2: Grounding multimodal large language models to the world.
\newblock \emph{arXiv preprint arXiv:2306.14824}.

\bibitem[{Qin et~al.(2023)Qin, Cai, Jin, Yan, Liang, Zhu, Lin, Han, Ding, Wang et~al.}]{qin2023webcpm}
Yujia Qin, Zihan Cai, Dian Jin, Lan Yan, Shihao Liang, Kunlun Zhu, Yankai Lin, Xu~Han, Ning Ding, Huadong Wang, et~al. 2023.
\newblock Webcpm: Interactive web search for chinese long-form question answering.
\newblock \emph{arXiv preprint arXiv:2305.06849}.

\bibitem[{Radford et~al.(2021)Radford, Kim, Hallacy, Ramesh, Goh, Agarwal, Sastry, Askell, Mishkin, Clark et~al.}]{radford2021learning}
Alec Radford, Jong~Wook Kim, Chris Hallacy, Aditya Ramesh, Gabriel Goh, Sandhini Agarwal, Girish Sastry, Amanda Askell, Pamela Mishkin, Jack Clark, et~al. 2021.
\newblock Learning transferable visual models from natural language supervision.
\newblock In \emph{International conference on machine learning}, pages 8748--8763. PMLR.

\bibitem[{Raffel et~al.(2019)Raffel, Shazeer, Roberts, Lee, Narang, Matena, Zhou, Li, and Liu}]{2019t5}
Colin Raffel, Noam Shazeer, Adam Roberts, Katherine Lee, Sharan Narang, Michael Matena, Yanqi Zhou, Wei Li, and Peter~J. Liu. 2019.
\newblock \href {https://arxiv.org/abs/1910.10683} {Exploring the limits of transfer learning with a unified text-to-text transformer}.
\newblock \emph{arXiv e-prints}.

\bibitem[{Rajpurkar et~al.(2016)Rajpurkar, Zhang, Lopyrev, and Liang}]{rajpurkar2016squad}
Pranav Rajpurkar, Jian Zhang, Konstantin Lopyrev, and Percy Liang. 2016.
\newblock Squad: 100,000+ questions for machine comprehension of text.
\newblock \emph{arXiv preprint arXiv:1606.05250}.

\bibitem[{Rawles et~al.(2023)Rawles, Li, Rodriguez, Riva, and Lillicrap}]{rawles2023android}
Christopher Rawles, Alice Li, Daniel Rodriguez, Oriana Riva, and Timothy Lillicrap. 2023.
\newblock Android in the wild: A large-scale dataset for android device control.
\newblock \emph{arXiv preprint arXiv:2307.10088}.

\bibitem[{Shaw et~al.(2023)Shaw, Joshi, Cohan, Berant, Pasupat, Hu, Khandelwal, Lee, and Toutanova}]{shaw2023pixels}
Peter Shaw, Mandar Joshi, James Cohan, Jonathan Berant, Panupong Pasupat, Hexiang Hu, Urvashi Khandelwal, Kenton Lee, and Kristina Toutanova. 2023.
\newblock From pixels to ui actions: Learning to follow instructions via graphical user interfaces.
\newblock \emph{arXiv preprint arXiv:2306.00245}.

\bibitem[{Shi et~al.(2017)Shi, Karpathy, Fan, Hernandez, and Liang}]{shi2017world}
Tianlin Shi, Andrej Karpathy, Linxi Fan, Jonathan Hernandez, and Percy Liang. 2017.
\newblock World of bits: An open-domain platform for web-based agents.
\newblock In \emph{International Conference on Machine Learning}, pages 3135--3144. PMLR.

\bibitem[{Singh et~al.(2019)Singh, Natarajan, Shah, Jiang, Chen, Batra, Parikh, and Rohrbach}]{singh2019towards}
Amanpreet Singh, Vivek Natarajan, Meet Shah, Yu~Jiang, Xinlei Chen, Dhruv Batra, Devi Parikh, and Marcus Rohrbach. 2019.
\newblock Towards vqa models that can read.
\newblock In \emph{Proceedings of the IEEE/CVF conference on computer vision and pattern recognition}, pages 8317--8326.

\bibitem[{Sun et~al.(2023{\natexlab{a}})Sun, Zhuang, Kong, Dai, and Zhang}]{sun2023adaplanner}
Haotian Sun, Yuchen Zhuang, Lingkai Kong, Bo~Dai, and Chao Zhang. 2023{\natexlab{a}}.
\newblock Adaplanner: Adaptive planning from feedback with language models.
\newblock \emph{arXiv preprint arXiv:2305.16653}.

\bibitem[{Sun et~al.(2022{\natexlab{a}})Sun, Chen, Chen, Dai, Zhu, and Yu}]{sun2022meta}
Liangtai Sun, Xingyu Chen, Lu~Chen, Tianle Dai, Zichen Zhu, and Kai Yu. 2022{\natexlab{a}}.
\newblock Meta-gui: Towards multi-modal conversational agents on mobile gui.
\newblock \emph{arXiv preprint arXiv:2205.11029}.

\bibitem[{Sun et~al.(2022{\natexlab{b}})Sun, Chen, Chen, Dai, Zhu, and Yu}]{sun2022metagui}
Liangtai Sun, Xingyu Chen, Lu~Chen, Tianle Dai, Zichen Zhu, and Kai Yu. 2022{\natexlab{b}}.
\newblock \href {https://arxiv.org/abs/2205.11029} {Meta-gui: Towards multi-modal conversational agents on mobile gui}.
\newblock \emph{Preprint}, arXiv:2205.11029.

\bibitem[{Sun et~al.(2023{\natexlab{b}})Sun, Shen, Cao, Liu, Li, Shen, Gan, Gui, Wang, Yang et~al.}]{llava-rlhf}
Zhiqing Sun, Sheng Shen, Shengcao Cao, Haotian Liu, Chunyuan Li, Yikang Shen, Chuang Gan, Liang-Yan Gui, Yu-Xiong Wang, Yiming Yang, et~al. 2023{\natexlab{b}}.
\newblock Aligning large multimodal models with factually augmented rlhf.
\newblock \emph{arXiv preprint arXiv:2309.14525}.

\bibitem[{Touvron et~al.(2023)Touvron, Lavril, Izacard, Martinet, Lachaux, Lacroix, Rozi{\`e}re, Goyal, Hambro, Azhar et~al.}]{touvron2023llama}
Hugo Touvron, Thibaut Lavril, Gautier Izacard, Xavier Martinet, Marie-Anne Lachaux, Timoth{\'e}e Lacroix, Baptiste Rozi{\`e}re, Naman Goyal, Eric Hambro, Faisal Azhar, et~al. 2023.
\newblock Llama: Open and efficient foundation language models.
\newblock \emph{arXiv preprint arXiv:2302.13971}.

\bibitem[{Venkatesh et~al.(2022)Venkatesh, Talukdar, and Narayanan}]{venkatesh2022ugif}
Sagar~Gubbi Venkatesh, Partha Talukdar, and Srini Narayanan. 2022.
\newblock Ugif: Ui grounded instruction following.
\newblock \emph{arXiv preprint arXiv:2211.07615}.

\bibitem[{Wu and Xie(2023)}]{vstar}
Penghao Wu and Saining Xie. 2023.
\newblock V*: Guided visual search as a core mechanism in multimodal llms.
\newblock \emph{arXiv preprint arXiv:2312.14135}.

\bibitem[{Xu et~al.(2021)Xu, Masling, Du, Campagna, Heck, Landay, and Lam}]{xu2021grounding}
Nancy Xu, Sam Masling, Michael Du, Giovanni Campagna, Larry Heck, James Landay, and Monica~S Lam. 2021.
\newblock Grounding open-domain instructions to automate web support tasks.
\newblock \emph{arXiv preprint arXiv:2103.16057}.

\bibitem[{Yang et~al.(2023)Yang, Liu, Han, Chen, Huang, Fu, and Yu}]{yang2023appagent}
Zhao Yang, Jiaxuan Liu, Yucheng Han, Xin Chen, Zebiao Huang, Bin Fu, and Gang Yu. 2023.
\newblock Appagent: Multimodal agents as smartphone users.
\newblock \emph{arXiv preprint arXiv:2312.13771}.

\bibitem[{Yao et~al.(2022)Yao, Chen, Yang, and Narasimhan}]{yao2022webshop}
Shunyu Yao, Howard Chen, John Yang, and Karthik Narasimhan. 2022.
\newblock Webshop: Towards scalable real-world web interaction with grounded language agents.
\newblock \emph{Advances in Neural Information Processing Systems}, 35:20744--20757.

\bibitem[{Yao et~al.(2024)Yao, Yu, Zhang, Wang, Cui, Zhu, Cai, Li, Zhao, He et~al.}]{yao2024minicpm}
Yuan Yao, Tianyu Yu, Ao~Zhang, Chongyi Wang, Junbo Cui, Hongji Zhu, Tianchi Cai, Haoyu Li, Weilin Zhao, Zhihui He, et~al. 2024.
\newblock Minicpm-v: A gpt-4v level mllm on your phone.
\newblock \emph{arXiv preprint arXiv:2408.01800}.

\bibitem[{Ye et~al.(2023)Ye, Hu, Xu, Ye, Yan, Xu, Li, Tian, Qian, Zhang, Jin, He, Lin, and Huang}]{ye2023ureader}
Jiabo Ye, Anwen Hu, Haiyang Xu, Qinghao Ye, Ming Yan, Guohai Xu, Chenliang Li, Junfeng Tian, Qi~Qian, Ji~Zhang, Qin Jin, Liang He, Xin~Alex Lin, and Fei Huang. 2023.
\newblock \href {https://arxiv.org/abs/2310.05126} {Ureader: Universal ocr-free visually-situated language understanding with multimodal large language model}.
\newblock \emph{Preprint}, arXiv:2310.05126.

\bibitem[{Yu et~al.(2023)Yu, Yao, Zhang, He, Han, Cui, Hu, Liu, Zheng, Sun et~al.}]{rlhf-v}
Tianyu Yu, Yuan Yao, Haoye Zhang, Taiwen He, Yifeng Han, Ganqu Cui, Jinyi Hu, Zhiyuan Liu, Hai-Tao Zheng, Maosong Sun, et~al. 2023.
\newblock Rlhf-v: Towards trustworthy mllms via behavior alignment from fine-grained correctional human feedback.
\newblock \emph{arXiv preprint arXiv:2312.00849}.

\bibitem[{Zhan and Zhang(2023)}]{zhan2023you}
Zhuosheng Zhan and Aston Zhang. 2023.
\newblock You only look at screens: Multimodal chain-of-action agents.
\newblock \emph{arXiv preprint arXiv:2309.11436}.

\bibitem[{Zhang et~al.(2024)Zhang, Hu, Khayatkhoei, Ilievski, and Sun}]{zhang2024exploring}
Jiarui Zhang, Jinyi Hu, Mahyar Khayatkhoei, Filip Ilievski, and Maosong Sun. 2024.
\newblock Exploring perceptual limitation of multimodal large language models.
\newblock \emph{arXiv preprint arXiv:2402.07384}.

\bibitem[{Zheng et~al.(2023)Zheng, Wang, Wang, and An}]{zheng2023synapse}
Longtao Zheng, Rundong Wang, Xinrun Wang, and Bo~An. 2023.
\newblock Synapse: Trajectory-as-exemplar prompting with memory for computer control.
\newblock In \emph{NeurIPS 2023 Foundation Models for Decision Making Workshop}.

\bibitem[{Zhou et~al.(2023)Zhou, Xu, Zhu, Zhou, Lo, Sridhar, Cheng, Bisk, Fried, Alon et~al.}]{zhou2023webarena}
Shuyan Zhou, Frank~F Xu, Hao Zhu, Xuhui Zhou, Robert Lo, Abishek Sridhar, Xianyi Cheng, Yonatan Bisk, Daniel Fried, Uri Alon, et~al. 2023.
\newblock Webarena: A realistic web environment for building autonomous agents.
\newblock \emph{arXiv preprint arXiv:2307.13854}.

\bibitem[{Zhu et~al.(2023)Zhu, Chen, Shen, Li, and Elhoseiny}]{zhu2023minigpt}
Deyao Zhu, Jun Chen, Xiaoqian Shen, Xiang Li, and Mohamed Elhoseiny. 2023.
\newblock Minigpt-4: Enhancing vision-language understanding with advanced large language models.
\newblock \emph{arXiv preprint arXiv:2304.10592}.

\bibitem[{Zhu et~al.(2024)Zhu, Zhu, Liu, Ou, Mou, and Tang}]{zhu2024llava}
Yichen Zhu, Minjie Zhu, Ning Liu, Zhicai Ou, Xiaofeng Mou, and Jian Tang. 2024.
\newblock Llava-phi: Efficient multi-modal assistant with small language model.
\newblock \emph{arXiv preprint arXiv:2401.02330}.

\end{thebibliography}

\appendix
\appendix

\section{Datasets}

\subsection{Action Space} 
\label{sec: actions}

\begin{figure}[tb]
  \centering
  \includegraphics[width=1.0\linewidth]{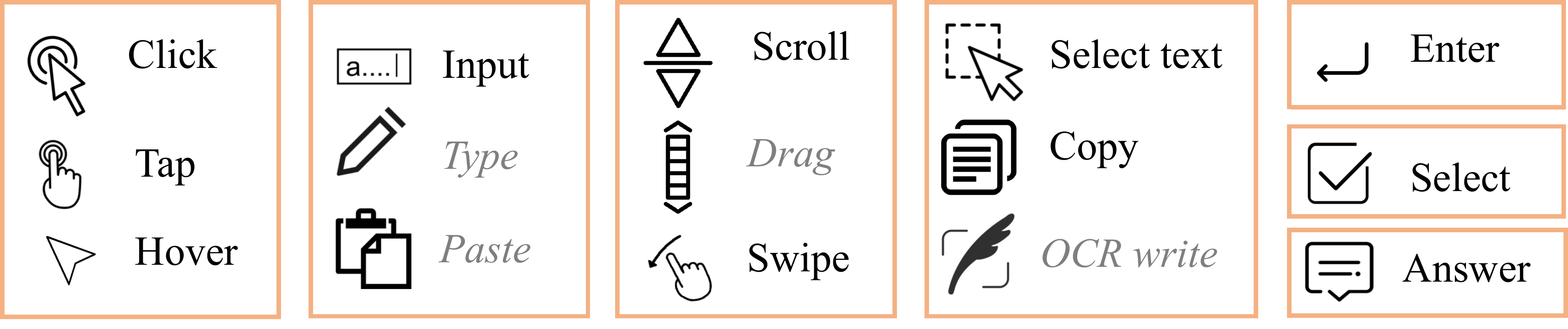}
  \caption{The actions in our action space. Our datasets utilize the actions displayed in black. The actions shown in \textcolor{gray}{gray} are not used, as their functions can be effectively replaced by other actions within the same group.}
  \label{fig: actions}
\end{figure}

In previous works, various definitions of the GUI agents' action space have been proposed. However, these approaches often suffer from one or more of the following issues: 
(1) The partitioning of actions is disorganized and lacks coherence.
(2) The action space fails to encompass all possible scenarios in GUI systems.
To address these limitations, our work introduces a unified action space for GUI systems. Specifically, we have designed seven distinct groups of actions tailored for task-oriented GUI agents, as depicted in Figure~\ref{fig: actions}:

(1) Pointing to something (e.g., buttons, icons): This action allows the agent to shift focus to specific elements within the GUI system, such as clicking on a search box or selecting empty spaces on websites. These actions include "click", "hover", and "tap".

(2) Inputting something: Agents can input relevant information into the GUI system, such as inputting search queries in a search box. These actions include "input", "type", and "paste".
    
(3) Browsing through pages: These actions enable the agent to navigate through pages within the GUI system, including page-up and page-down functionalities. These actions include "scroll", "drag", and "swipe".
    
(4) Logging information: Agents can log key information that aids in accomplishing the task, such as writing texts or copying essential paragraphs. These actions include "select text", "copy", and "write (by OCR)".

(5)  Submitting (tables, search text): Agents can submit tables or the search context within the GUI system. We use the "enter" action to achieve this target.
    
(6) Selecting a value: This action allows the agent to choose a specific value from a drop-down menu. We use the "select" action to achieve this target.
    
(7) End of task: Agents can signal the completion of the task or respond (e.g., an answer, task completion status, or indication of task impossibility). We use the "answer" action to achieve this target.

\paragraph{Actions Conversation from AITW dataset.} During the contribution of our GUIAct-smartphone dataset, we use a subset of the AITW \citep{rawles2023android} dataset, a large smartphone navigation data with General/Install/Google-Apps/Single/WebShopping tags. We use a subset of the partition with the "General" tag. To keep the consistency of our action space, we convert their actions: 
(1) dual-point gesture: we split the "dual-point gesture" actions into "tap" and "swipe" by the distance of the touch point and the left point. Then, we use the touch point as the "point" parameter of the "tap" action and use both points as the "dual-point" parameter of the "swipe" action.

(2) type: we rename the "type" action as the "input" action, and reserve the "text" parameter of "type" as the "text" parameter of "input".

(3) enter: we reserve the "enter" action.

(4) go back/home: we reflect on these two actions to "tap" actions. We find the position of the two buttons on the bottom navigation bar as the "point" parameter of the "tap“ action.

(5) task complete/impossible: we convert these two actions to "answer" actions, and the "task complete" and "task impossible" are the "text" parameters of the "answer" action.

\paragraph{The Format of Actions.} Two types of formats need to be considered for actions: parse format and position format. We need a structured text format to extract actions from GUI agents' natural language responses, referred to as the parse format. Also,  we need an absolute or related format to express the position information of the actions, referred to as the position format.

\begin{figure}[tb]
  \centering
  \includegraphics[width=1.0\linewidth]{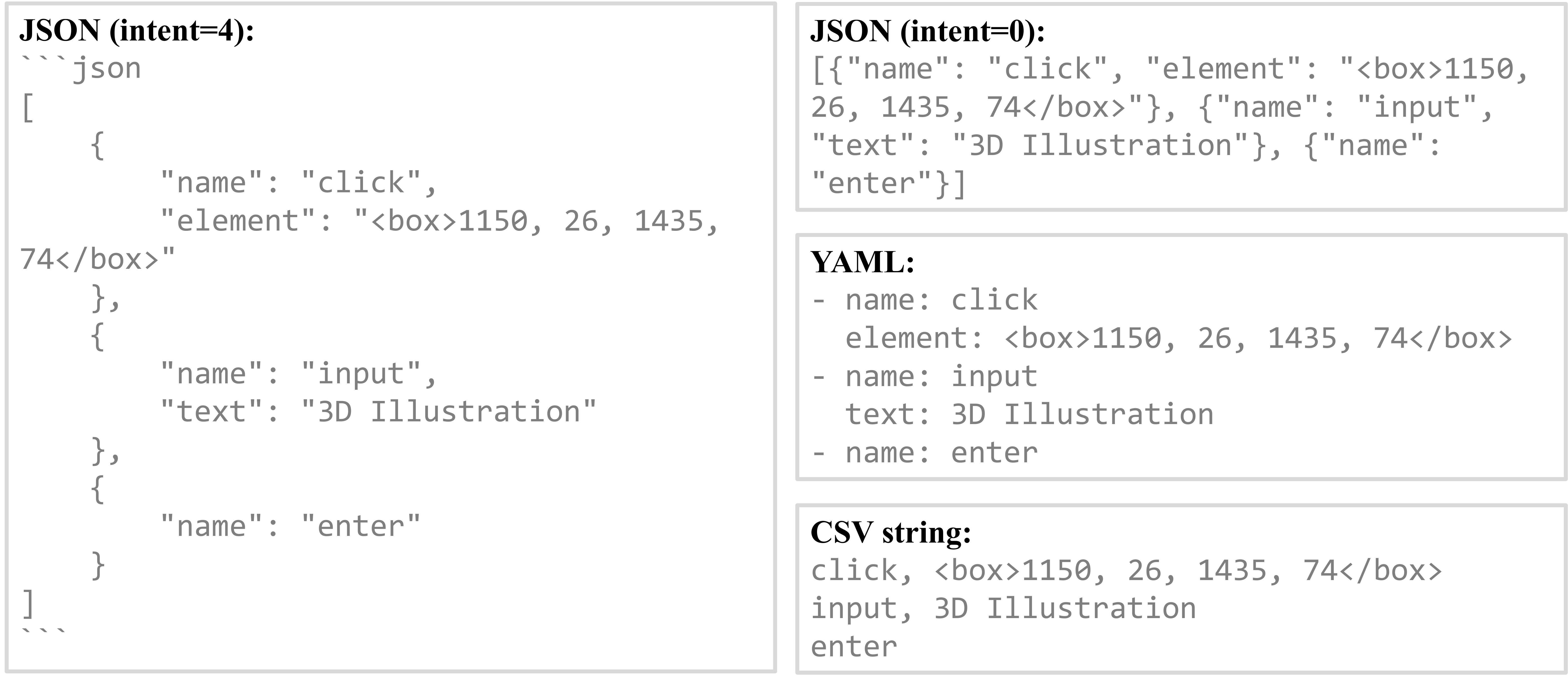}
  \caption{The four different action parse formats.}
  \label{fig: parse format}
\end{figure}

As for \textbf{parse format}, we try several formats such as JSON and YAML, as shown in Figure~\ref{fig: parse format}. In our experiments, we utilize JSON format as the action parse format for Qwen-VL and Fuyu-8B. For MiniCPM-V, we opt for the CSV format, which closely resembles natural language conventions.

As for \textbf{position format}, we use different position formats based on the models' preferences: absolute position format for Fuyu-8B and related position format for Qwen-VL. We also have a variant of related position formats, used for MiniCPM-V, which is scaling the related value to [0,1000) as integers (e.g., "<box>481 565 506 592</box>").  We show the position formats and the four parameters in Table~\ref{tab: position formats.}.

\begin{table*}[tb]
    \centering
    \small
    \renewcommand{\arraystretch}{1.1}
    \setlength\tabcolsep{1mm}{
    \begin{tabular}{llll}
        \toprule
        \textbf{Actions} & \textbf{Param.} & \textbf{Absolute} & \textbf{Related}\\
        \midrule
        \textbf{Click, Hover, Select} & \textbf{element} & <box>657, 434, 691, 455</box> &  <box>0.481, 0.565, 0.506, 0.592</box> \\
        \textbf{Tap} & \textbf{point} & <point>362, 1412</point> & <point>0.503, 0.981</point> \\
        \multirow{2}{*}{\textbf{Swipe, Select\_Text}} & \multirow{2}{*}{\textbf{dual points}} & from: <point>884, 236</point> & from: <point>0.730, 0.311</point> \\
        & & to: <point>958, 233</point> & to: <point>0.791, 0.307</point> \\
        \textbf{Scroll} & \textbf{scroll} & down 496 right 0 & down 0.65 right 0 \\
        \bottomrule
    \end{tabular}}
    \caption{The position and distance formats. An element's position is composed of its left-top point and right-down point and can be expressed as <box>x1, y1, x2, y2</box>. A point's position is composed of its x and y coordinates, expressed as <point>x, y</point>. The position of dual points is composed of its start point and end point. The scroll parameter's distance information uses pixel values. There are "down" and "right" two parameters to signal y dim and x dim respectively. If the action is scrolling up, the value of the "down" parameter is negative.}
    \label{tab: position formats.}
\end{table*}

\subsection{LLM-Auto Annotation}
\label{sec: prompts}
We use LLMs to help our contribution processing for GUIAct (\todot{single-step and multi-step website navigation}), and GUIChat datasets.

\subsubsection{GUIAct (\todot{single-step website navigation})}

\begin{figure*}[tb]
  \centering
  \includegraphics[width=1.0\linewidth]{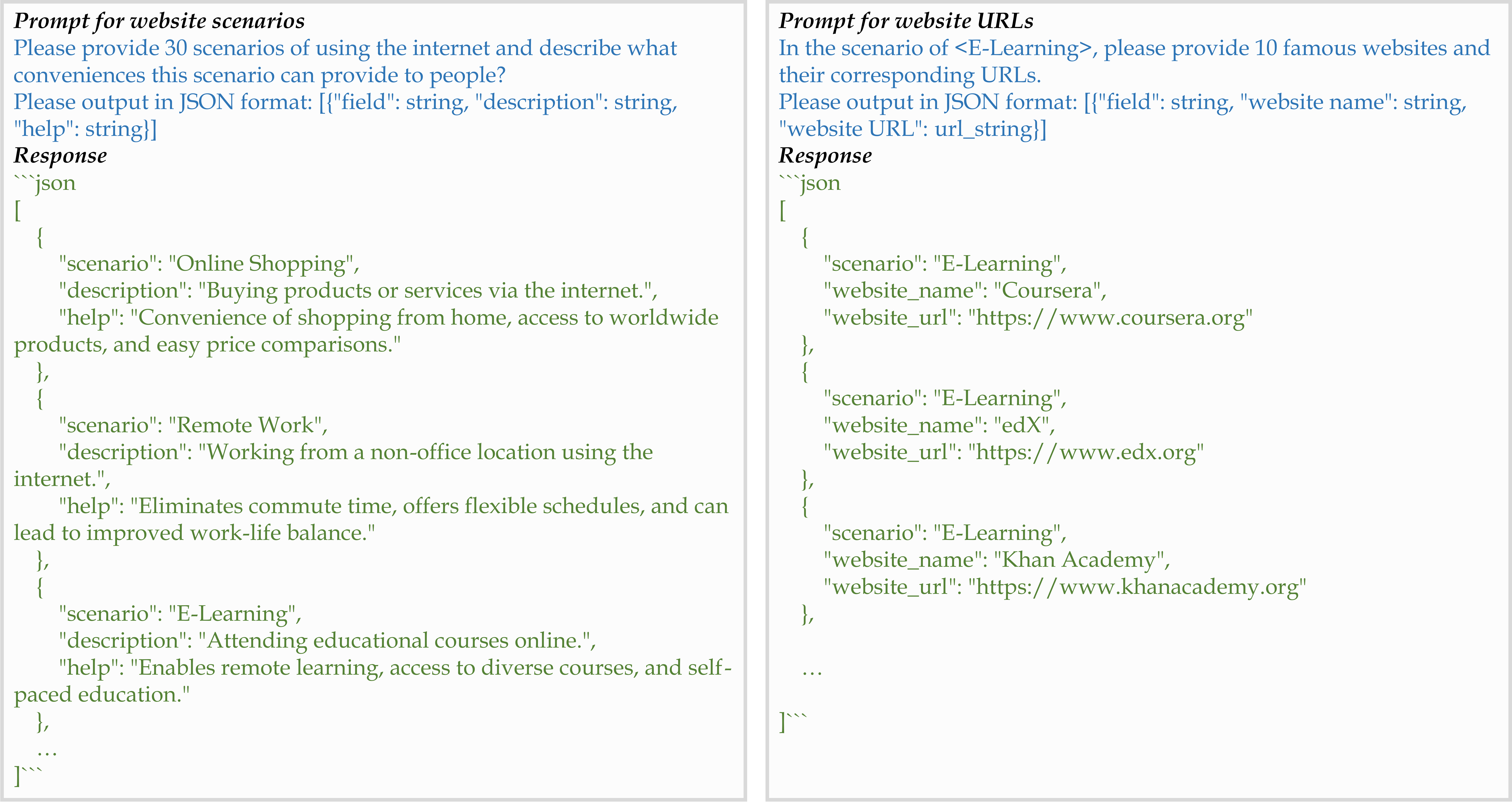}
  \caption{The prompts for acquiring website scenarios and website URLs.}
  \label{fig:guiact-single-prompt1}
\end{figure*}

\paragraph{Website Selection.} We use GPT-4 to gather website scenarios and website URLs, the prompts are shown in Figure~\ref{fig:guiact-single-prompt1}. 

\begin{figure*}[tb]
  \centering
  \includegraphics[width=1.0\linewidth]{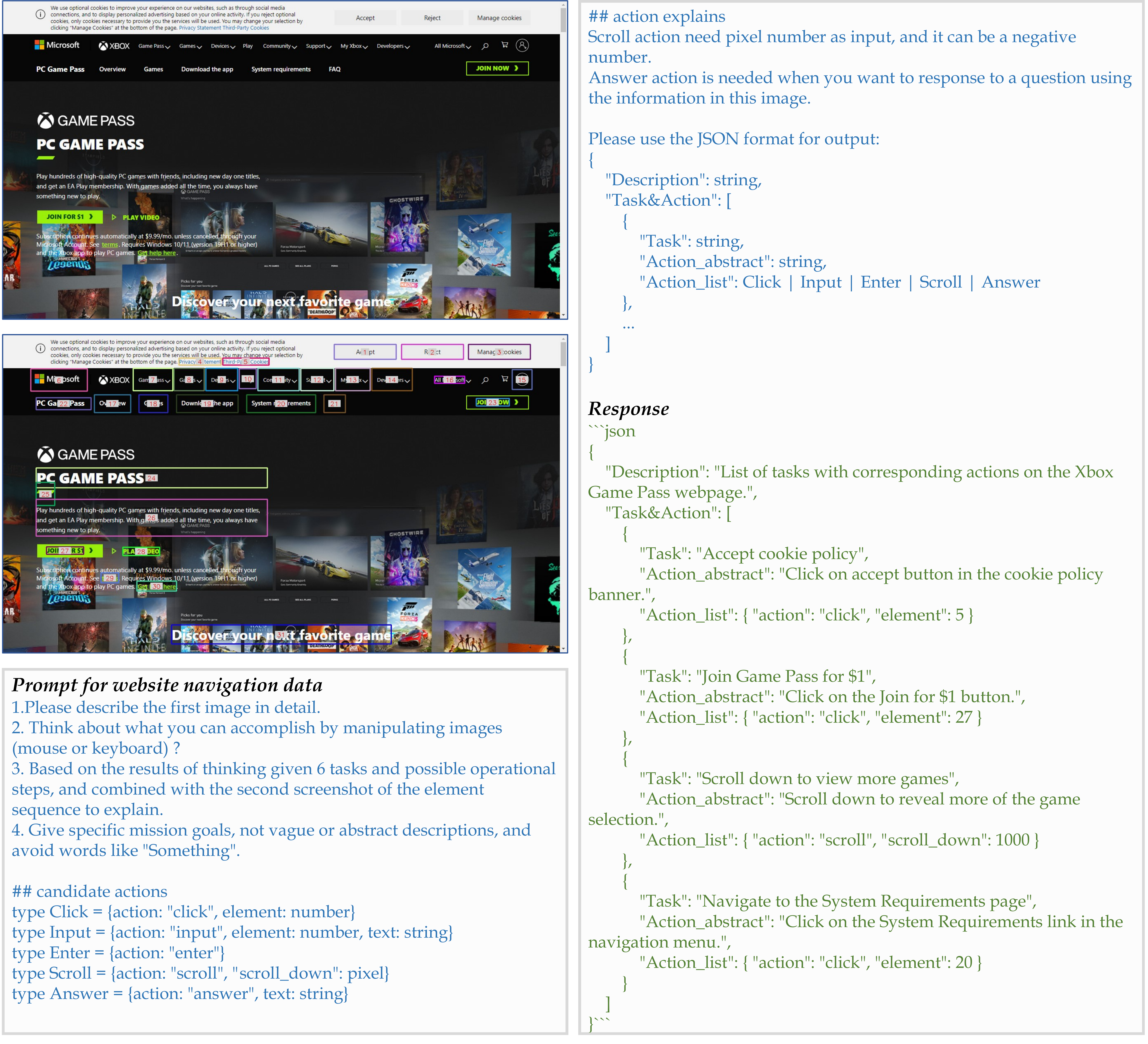}
  \caption{The prompts and images for acquiring question-action pairs by GPT-4V.}
  \label{fig:guiact-single-prompt2}
\end{figure*}

\paragraph{LLM-Auto Annotation.} We use GPT-4V to obtain instruction-action pairs for each web screenshot. To help GPT-4V comprehend both the content and the index of the elements, we provide two images for each request. As shown in Figure~\ref{fig:guiact-single-prompt2}, one image is the origin screenshot without any mask box, and the other image is a revised screenshot with an element index and mask box on each element. Due to that, GPT-4V can get full information on the website viewport in the first image and get the element ID information in the second image. 

\subsubsection{GUIAct (\todot{multi-step website navigation})}

\begin{figure*}[tb]
  \centering
  \includegraphics[width=1.0\linewidth]{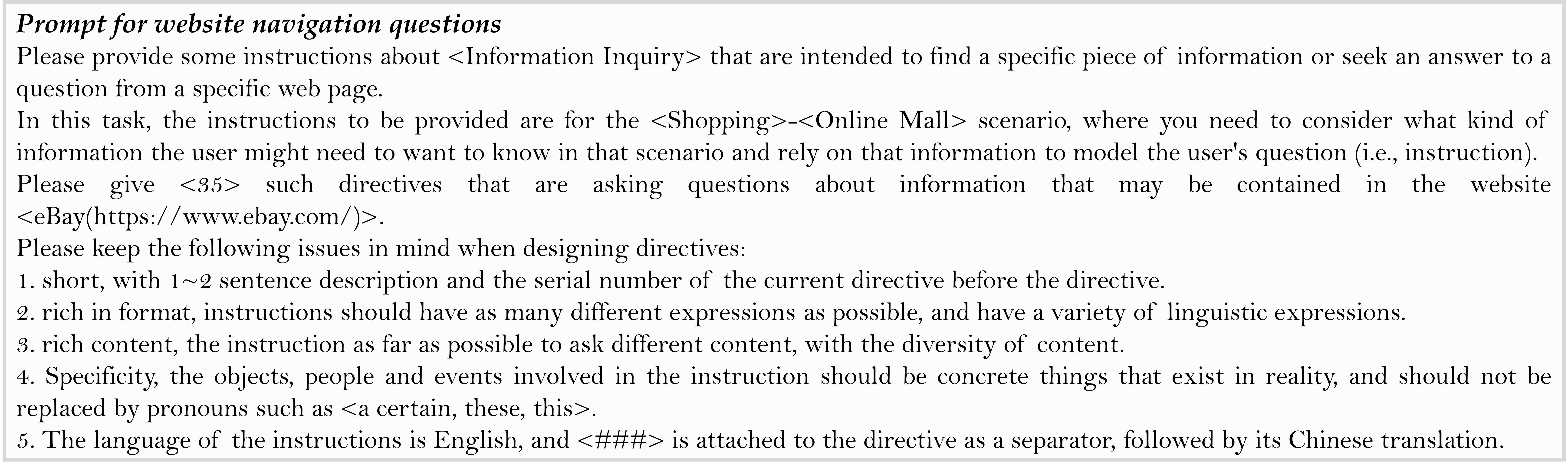}
  \caption{The prompt template for acquiring questions about information inquiry on a designated website.}
  \label{fig:guiact-multi-prompt}
\end{figure*}

\paragraph{Questions Acquisition.} We use LLMs (GPT-3.5 and Claude2) to generate English and Chinese questions for English and Chinese websites respectively. These questions all have special answers due to the information on the websites. The prompt template is shown in Figure~\ref{fig:guiact-multi-prompt}.

\subsubsection{\todot{GUIChat}}

\begin{figure*}[ht]
  \centering
  \includegraphics[width=1.0\linewidth]{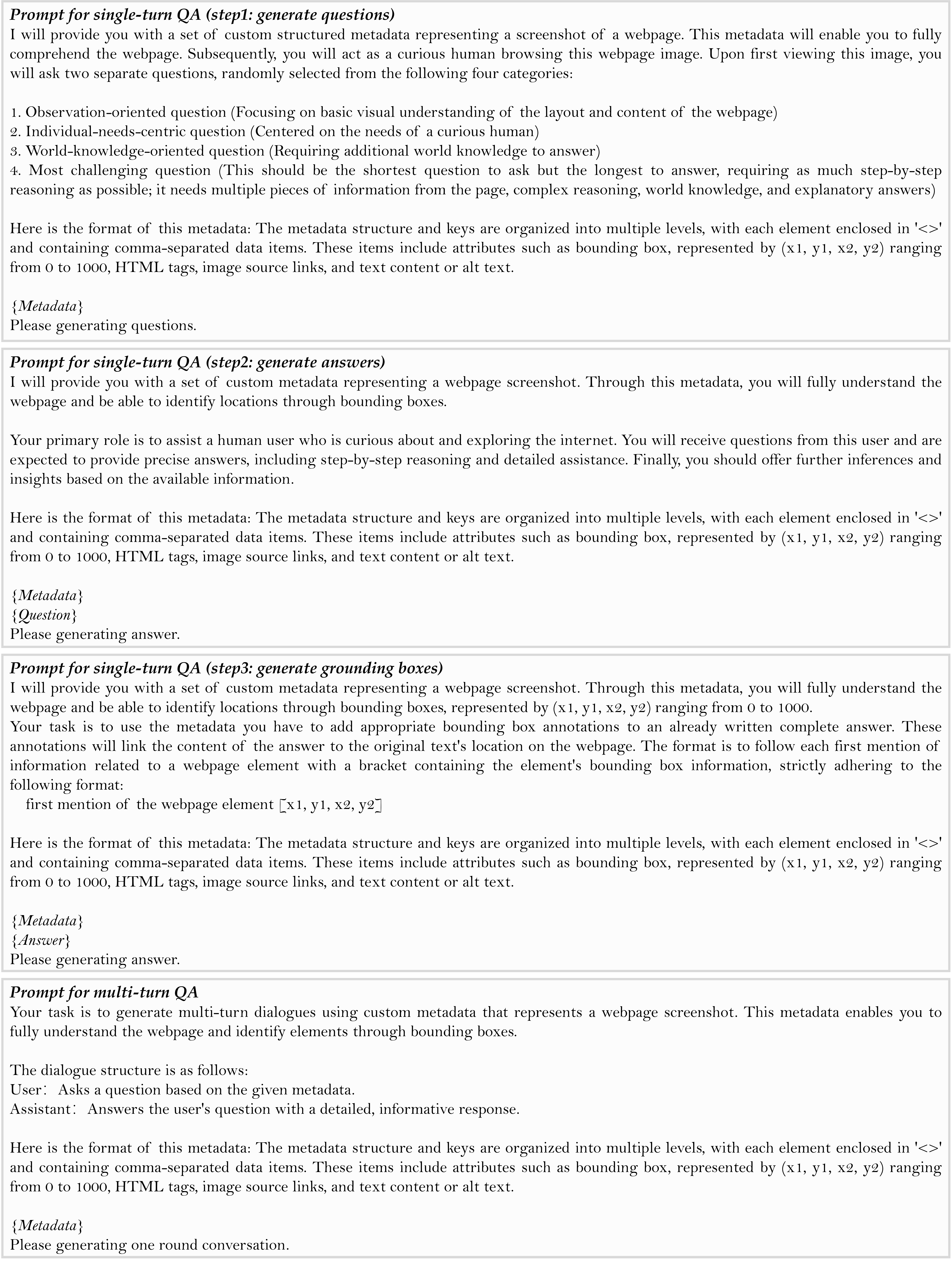}
  \caption{The prompts for generating the GUIChat dataset.}
  \label{fig:guichat-prompt}
\end{figure*}

\paragraph{Question-Answer Pair Generation.} We use GPT-4 and employ a pipeline for generating both single-turn and multi-turn question-and-answer pairs. For single-turn question-and-answer generation, the process begins with generating 4 types of questions, followed by generating answers and answers with grounding. Multi-turn question-and-answer generation involves generating one round of questions and answers per turn based on the history of the conversation, with a maximum of five rounds. The prompts are shown in Figure~\ref{fig:guichat-prompt}.

\subsection{Crowd-sourced Annotation}
\label{sec: handbook}
We hire humans to help us check or annotate the GUIAct (\todot{single-step and multi-step website navigation}), and GUIChat datasets.

\subsubsection{GUIAct (\todot{single-step website navigation})}
\paragraph{Data checking by human.} 
\todot{When we obtained auto-generated data from GPT-4V, we randomly sampled about 400 instances and discovered numerous errors, resulting in an overall accuracy of only about 55\%. To address this, we contracted a professional data annotation company. We paid 0.6 RMB per instance to verify the LLM’s automatic annotations. With each instance taking approximately 40 seconds to check, the hourly wage for the annotators was 54 RMB. The entire annotation phase lasts for two months, from November 2023 to January 2024.}

\todot{We have about 78k instances for review. Errors that were easy to correct, such as incorrect element IDs or action names, were adjusted accordingly. For more complex errors, such as unfiltered web pages, overall element displacement, unreasonable instructions, or mismatches between thoughts and actions, the annotators deleted the problematic samples. As a result, we ended up with 67k samples and improved the accuracy from 55\% to 92\%.}

\todot{The checking process was divided into three batches: 7,548 samples in the first batch, 30,998 in the second, and 40,000 in the third. Each batch underwent a three-step review process: initial checking by data annotators, subsequent review by a company reviewer, and final acceptance by us. We sampled 10\% of the first batch and 2\% of each of the subsequent batches for quality control. Each batch was also subdivided into smaller batches (1,000 or 5,000 samples). If the accuracy in any sub-batch fell below 90\%, annotators were required to recheck all data in that sub-batch until accuracy exceeded 90\%. Overall, we achieved a total accuracy of approximately 92\%.}

\subsubsection{GUIAct (\todot{multi-step website navigation})}
\begin{figure*}[tb]
  \centering
  \includegraphics[width=1.0\linewidth]{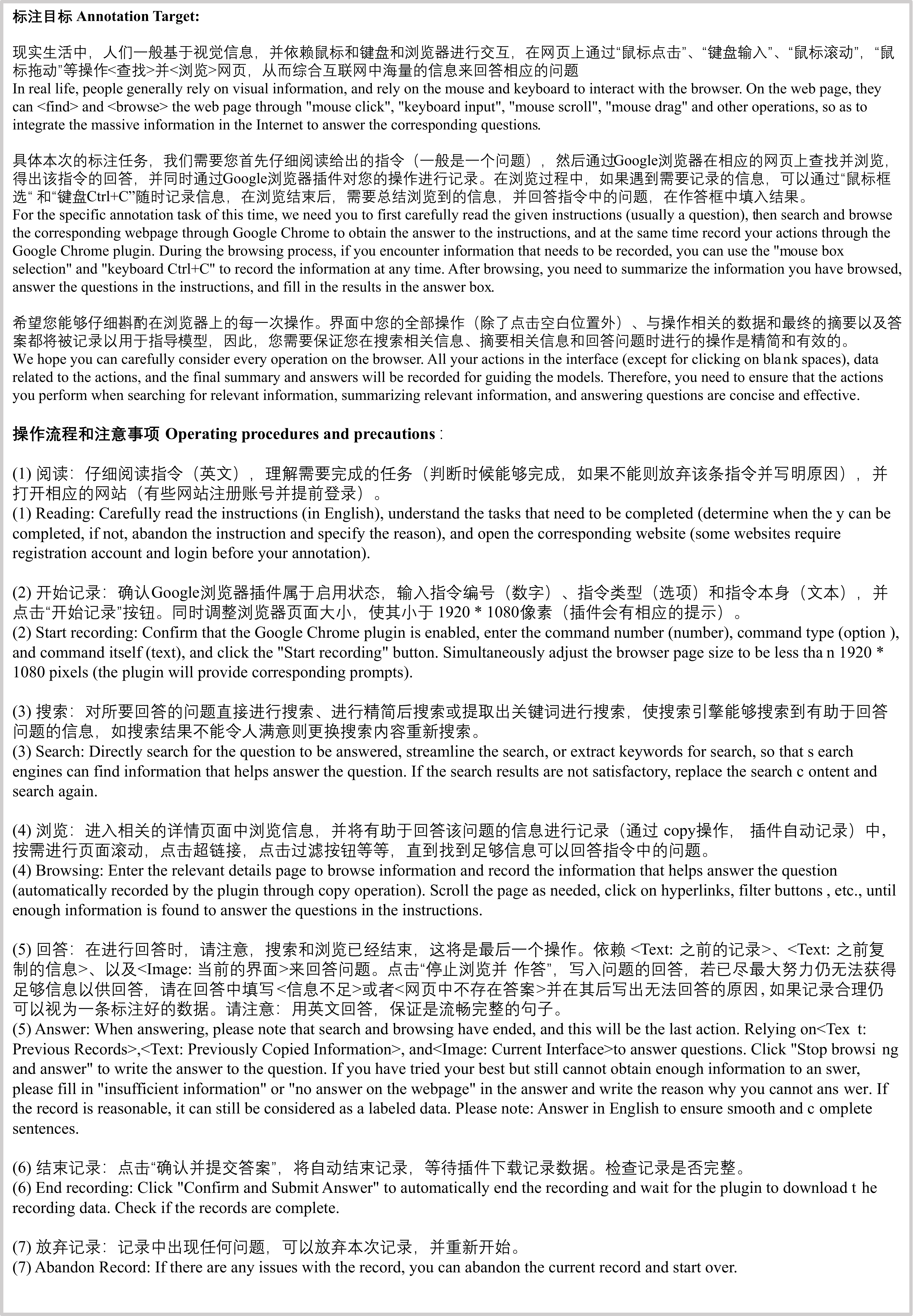}
  \caption{The main content of our data annotation handbook for GUIAct (web-multi).}
  \label{fig: handbook}
\end{figure*}

\paragraph{Crowd-sourced Annotation.} 
\todot{We have a data annotation handbook for the annotation task for GUIAct (web-multi) data. We show the main content in Figure~\ref{fig: handbook}. We pay 5 RMB per item to annotate these data, and the annotation efficiency is approximately 5 minutes per item. Therefore, the hourly salary we pay the annotator is 60 RMB. The entire annotation phase lasts for four months, from September 2023 to January 2024.}


\subsubsection{\todot{GUIChat}}

\paragraph{Question-Answer Pair Evaluation.} \todot{In order to ensure the quality of GUIChat dataset, we evaluate by humans. The metrics we designed include "Helpfulness", "Honesty", "Harmlessness", and "Image-text correlation" four aspects: (1) Helpfulness means that the answer addresses the user’s question or request comprehensively; (2) Honesty means that the answer is factually accurate and does not have false or misleading meanings, and has roughly accurate bounding boxes; (3) Harmlessness means that the screenshots, questions, and answers do not contain offensive, discriminatory, or harmful content; (4) Image-text correlation means that the question can not be answered accurately without the screenshot. We will give a score for each aspect. If annotators think that a question-answer pair meets the requirements of a certain aspect, they will give it 1 point, otherwise 0 point. There are 44k single-turn and 6k multi-turn dialogue instructions in our GUIChat dataset. \todop{We have randomly sampled 500 question-answer pairs and evaluated them by humans. The average scores for the four aspects are $0.954$, $0.814$, $0.958$, and $0.886$ respectively.}
These results demonstrate that the quality of our GUIChat dataset is sufficient for training purposes. Most errors stem from screenshots of unloaded websites. 
}
\section{Experiments}
\label{sec: experiments}

\subsection{Training Details}
\label{sec: training details}

\paragraph{Dataset Partitions.} Our GUICourse includes three datasets: GUIEnv, GUIAct, and GUIChat. 

(1) GUIEnv has 10M samples for pre-training (GUIEnv-global) and 0.7M samples for SFT (GUIEnv-local). Meanwhile, we split 1.8k samples in GUIEnv-local as test data for OCR and grounding abilities evaluation. 

(2) GUIAct is composed of three partitions: GUIAct (web-single), GUIAct (web-multi), and GUIAct (smartphone). GUIAct (web-single) is a dataset with single-step actions in website scenarios, which has about 67k instructions automatically generated by 13k screenshots. GUIAct (web-multi) is a dataset with multi-step actions and 5,696 complex instructions, which are annotated by humans. Also, we simply 9,157 instructions from AITW for smartphone scenarios, referred to as GUIAct (smartphone). There are a total of 67k, 44k, and 67k samples for the three partitions, and we split about 1.4k, 1k, and 2k samples as testing data. The residual samples are used for SFT training. 

(3) GUIChat has 44k single-turn QA pairs and 6k multi-turn dialogues. We use all the samples for SFT.

\paragraph{Training Stages.} Our training pipeline can be split into two stages due to the training targets. In the pre-training stage, we mainly improve the OCR and text-grounding abilities of VLMs, especially in complex website scenarios. In the SFT training stage, we mainly enhance the VLMs' knowledge about GUI systems, such as the function of elements and control actions, as well as the conversation abilities.

We only execute the whole training pipeline based on MiniCPM-V. For other used VLMs (such as Qwen-VL and Fuyu-8B), for which we can't acquire their pre-training data and details, we only execute the SFT training stage. 
(1) Pre-training stage: we mix GUIEnv-global data in the pre-training data and then execute the training. 
(2) SFT stage: we merge the data from three partitions as SFT data: 10\% of the GUIEnv-local, all the data of GUIAct, and all the data of GUIChat. The learning rate of MiniCPM-V is 5e-7, while the learning rate of Qwen-VL and Fuyu-8B is 1e-5. All the learning scheduler types are "cosine".

\paragraph{Resources Cost.} For each agent, We use 1 \(\times\) GPU A100 for inference, 8 \(\times\) GPU A100 for SFT, and 128 \(\times\) GPU A100 for pre-training. As for MiniCPM-V (3.1B), it takes about 2 days for the whole pre-training stage using the GUIEnv dataset and about 8 hours for the SFT stage using GUIAct and GUIChat datasets with 3 epochs. Moreover, it takes several hours for the evaluation. As for Qwen-VL (9.6B) and Fuyu-8B (9.4B), we only do SFT and inference processes. It takes about 1 day for training with the same SFT data. 



\subsection{The Evaluation of Actions.}
\label{sec: evaluation details}

\paragraph{Type Exact Match Score.} We use Type Exact Match Score (Type EM) to calculate the accuracy for predicting the names of actions, and not care about the parameters of the actions. The Type EM score is also called action type match \citep{zhan2023you} or intend match \citep{lù2024weblinx} in other works. 

\begin{table*}[!t]
    \centering
    \small
    \setlength\tabcolsep{3mm}{
    \begin{tabular}{lllll}
        \toprule
        \textbf{Action} & \textbf{Param.} & \textbf{Examples} & \textbf{Score} & \textbf{Is Success} \\
        \midrule
        \textbf{Click} & element & <box>32, 20, 56, 44</box> & IoU  & EM(el.)\\
        \textbf{Tap} & point & <point>183, 505</point> & distance & distance < 0.14 \\
        \textbf{Hover} & element & <box>299, 118, 347, 166</box> & IoU  & EM(el.)\\
        \textbf{Input} & text & "French butter cake" & F1 & F1 > 0.5 \\
        \textbf{Scroll} & pixel & down 500 right 0 & direction & direction \\
        \textbf{Swipe} & dual points & from <point1> to <point2> & direction & direction\\
        \textbf{Select Text} & dual points & from <point1> to <point2> & IoU & IoU > 0.5 \\
        \textbf{Copy} & - & - & EM(type) & EM(type) \\
        \textbf{Enter} & - & - & EM(type) & EM(type) \\
        \textbf{Select} & element, text & the <element> value to "Female" & IoU + F1 & EM(el.) + F1\\
        \textbf{Answer} & text & "On April 16, 2021." & F1 & F1 > 0.5 \\
        \bottomrule
    \end{tabular}}
    \caption{The metrics of different actions.}
    \label{tab:action evaluaton}
\end{table*}

\begin{figure*}[tb]
  \centering
  \includegraphics[width=0.9\linewidth]{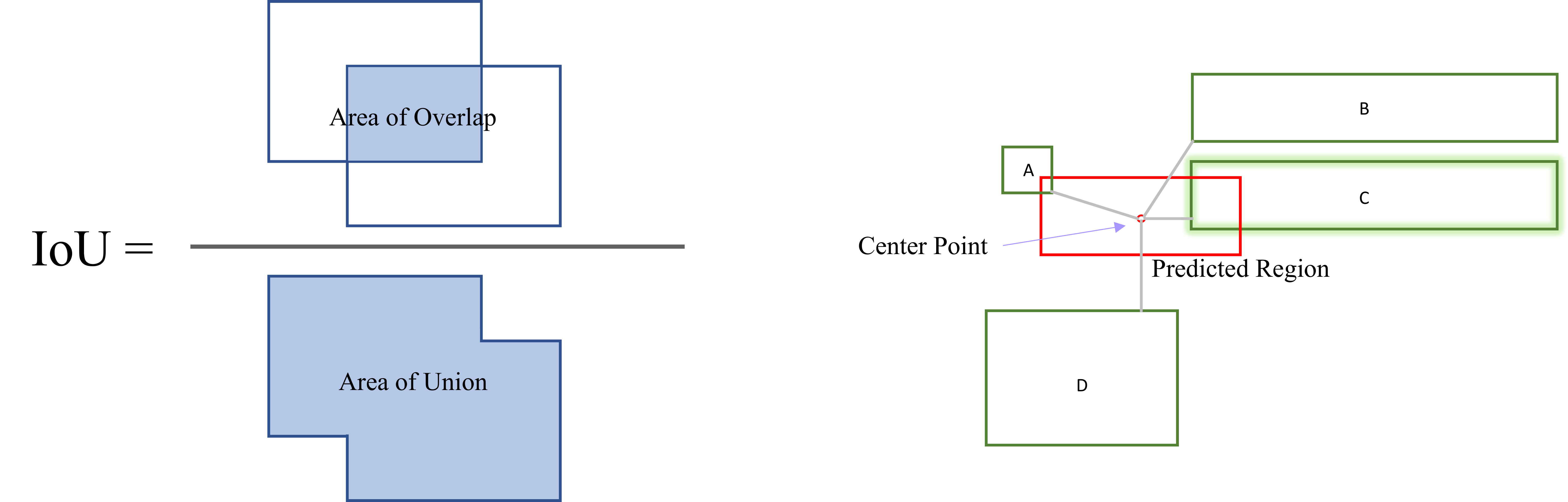}
  \caption{The IoU score and the element attach method. The left partition is the metrics of IoU scores and the right partition is the sketch of the element attach method. We first calculate the center point of the predicted element, and then calculate the distances (the gray dashed line) from this point to all the candidate elements. We select the element with the minimum distance as the final predicted element (the element "C" in this figure).}
  \label{fig: iou-attach}
\end{figure*}

\paragraph{Action Score and Success Rate.} We calculate action scores and success rates for every action, as shown in Figure~\ref{tab:action evaluaton}. The action scores are calculated similarity between the predicted actions and the golden actions. However, the action success rates judge if the predicted actions achieve the same function as the golden actions. We design different rules to calculate the action scores and success rates for various actions in our action space. 

(1) "click" and "hover": For evaluating the two actions, we employ the Intersection over Union (IoU) score of their respective element regions (referred to as boxes) as their action scores. To determine the action success rate, we compare the predicted elements' IDs with the golden elements' IDs using an exact match criterion. It is worth noting that we associate the predicted region with the element having the minimum distance, as illustrated in Figure~\ref{fig: iou-attach}. Generally, the success rates of "click" and "hover" actions (referred to as Cli.Acc) tend to be higher than their corresponding action scores.
       
(2) "tap": For the "tap" action, there are no candidate or golden regions available in the datasets. Instead, we determine the success of the action by calculating the distance between the predicted point and the golden point using the related position format. If the distance is less than 0.14, the action is considered successful, and the action score is calculated as $1 - \frac{distance}{0.14}$. Otherwise, the action score and success rate are set to $0.0$.
    
(3) "input" and "answer": We utilize the F1 score as the action score. If the action score exceeds $0.5$, we consider the action to be successful.

(4) "scroll" and "swipe": We convert the parameters of these actions into four directions: up, down, right, and left. If the predicted and golden results have consistent directions, the action is deemed successful, and the action score is set to $1.0$. Otherwise, the action score and success rate are set to $0.0$.

(5) "copy" and "enter": There are no parameters in both actions. The action scores and the success rates are equal to the Type EM scores.
    
(6) “select\_text”: This action involves selecting text within a swiped region. We evaluate the action score using the IoU metric. If the IoU score exceeds $0.5$, we consider the action to be successful.
    
(7) “select”: there are two parameters (element and text) in this action. We use the IoU and F1 score for its element parameter and text parameter respectively. We calculate the average of the two scores as the action score. The action is considered successful only if both parameters meet the success criteria.

\paragraph{Step Success Rate.} In our approach, we encounter scenarios where multiple actions need to be performed in a single step. For instance, based on a given screenshot, the required actions could be a combination of "click," "input," and "enter." Therefore, our datasets necessitate models that can predict one or multiple actions in a single step. To Evaluate the performance of a single step, \emph{(1)} we first compare the top-n predicted actions with the golden actions, where $n$ represents the number of golden actions. If the number of predicted actions is less than $n$, the scores of missing actions are set to $0.0$. \emph{(2)} Then, we calculate the average action score for each action that is being compared, providing an overall action score. \emph{(3)} Finally, we calculate the step Success Rate (StepSR). We consider a step successful only if all the actions within that step are successfully executed.

\subsection{Case Study}
\label{sec: android}

\begin{figure*}[tb]
  \centering
  \includegraphics[width=1.0\linewidth]{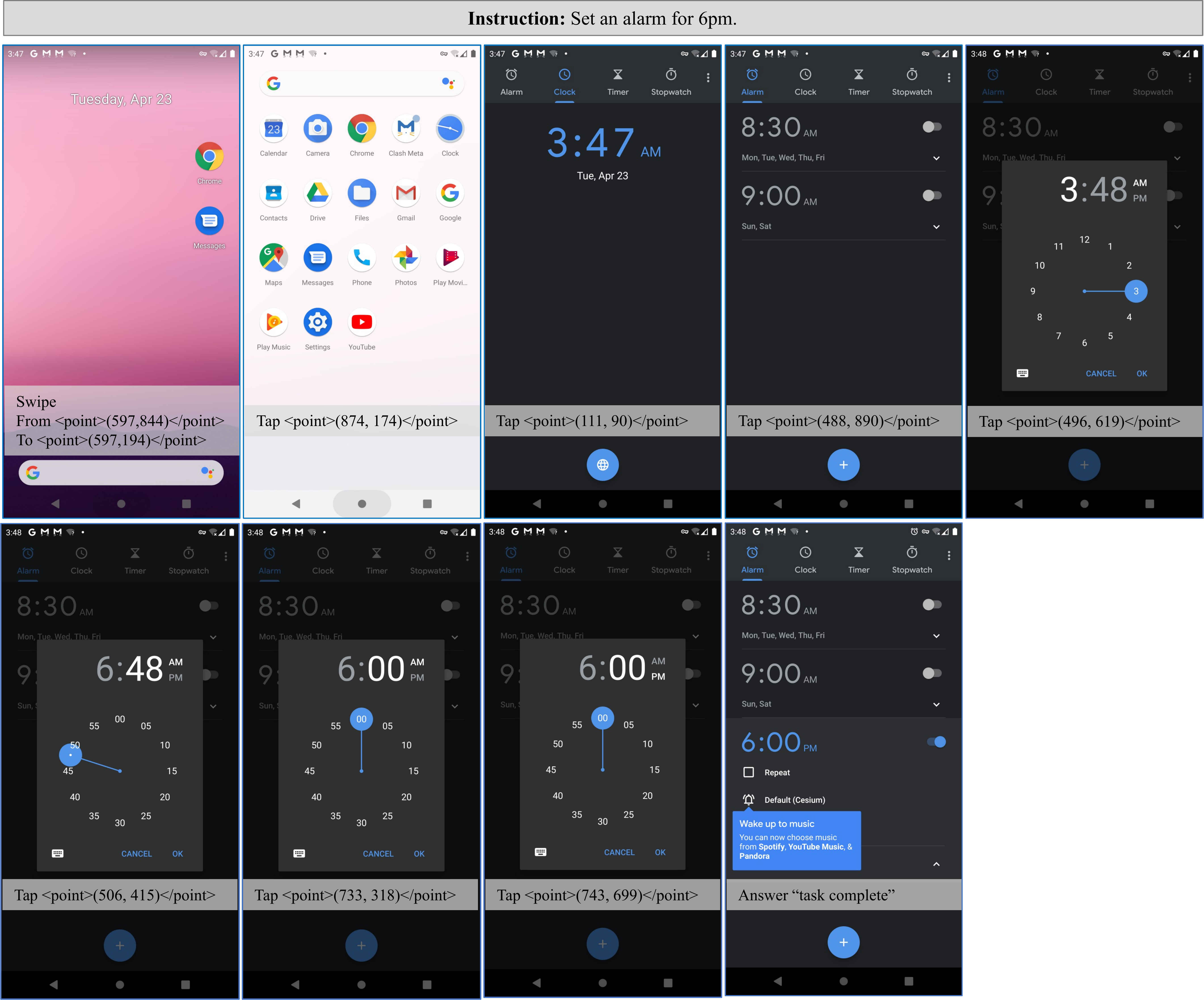}
  \caption{An example of our GUI agent executing the instruction with multiple steps in the simulated smartphone environment.}
  \label{fig: multistep-ins}
\end{figure*}

We show a long-chain example executing in the simulated smartphone environment in Figure~\ref{fig: multistep-ins}.

\subsection{External Experiments}
\label{sec: Larger LLM}

\subsubsection{\todot{MiniCPM-GUI with Larger LM}}
\todot{We also train a larger GUI agent (MiniCPM-GUI with larger LM), similar in size to Qwen-VL, by combining GUIEnv data with its pre-training data and fine-tuning it using the GUIAct and GUIChat datasets. The new agent has better performance on our test data and outperforms SeeClick on Mind2Web, demonstrating the effectiveness of our approach when scaling up model size and incorporating comprehensive datasets. The results are shown in Figure~\ref{tab: mind2web-additional-results} and Figure~\ref{tab: guiact-additional-results}.}

\subsubsection{\todot{GUIAgent with GUIChat Data.}}
\todot{The main target of GUIChat data is to improve the dialogue ability of GUI agents. We have experiments that comparing the results using "GUIAct + GUIChat" datasets and "GUIAct" only based on MiniCPM-V and QWen-VL. As shown in Figure~\ref{tab: guiact-additional-results}, the results are similar, which means the improvement of agents’ conversation ability has little impact on the action execution ability (here is the GUI navigation ability).}

\begin{table}[tb]
  \small
  \centering
  \setlength\tabcolsep{0.6mm}{
  \begin{tabular}{lccccccc}
    \toprule
    \multirow{2}{*}{\textbf{Agents}} & \multicolumn{2}{c}{Cross-Task} & \multicolumn{2}{c}{Cross-Website} & \multicolumn{2}{c}{Cross-Domain}\\
    & \scriptsize Ele.Acc & \scriptsize StepSR & \scriptsize Ele.Acc & \scriptsize StepSR & \scriptsize Ele.Acc & \scriptsize StepSR \\
    \midrule
    SeeClick & 28.3 & 25.5 & 21.4 & 16.4 & 23.2 & 20.8 \\
    \midrule
    Qwen-GUI & 27.9 & 24.4 & 19.3 & 15.6 & 20.5 & 17.5\\
    Fuyu-GUI & 19.1 & 15.6 & 13.9 & 12.2 & 14.2 & 11.7 \\
    \midrule
    MiniCPM-GUI & 23.8 & 20.8 & 20.3 & 17.3 & 17.9 & 14.6 \\
    \todot{+ Larger LM} & \textbf{36.9} & \textbf{33.2} & \textbf{34.7} & \textbf{29.7} & \textbf{35.7} & \textbf{31.6} \\
    \bottomrule
  \end{tabular}}
  \caption{Results of our GUI agents on Mind2Web. \todot{With a larger LM, a similar size as SeeClick, MiniCPM-GUI outperforms it.}}
  \label{tab: mind2web-additional-results}
\end{table} 

\begin{table*}[tb]
  \small
  \centering
  \setlength\tabcolsep{0.7mm}{
  \begin{tabular}{lcccccccccc}
    \toprule
    \multirow{2}{*}{\textbf{Agent}} & \multicolumn{3}{c}{Web-Single} & \multicolumn{3}{c}{Web-Multi} & \multicolumn{3}{c}{Smartphone} & Mean \\
    & Type EM & Cli.Acc & StepSR & Type EM & Cli.Acc & StepSR & Type EM & Cli.Acc & StepSR & StepSR\\
    \midrule
    \todot{GPT-4o-mini*} & 81.0 & 62.3 & 57.0 & 22.0 & 10.0 & 17.0 & 53.0 & 13.0 & 22.0 & 32.0 \\
    \midrule
    Fuyu-GUI & 90.1 & 67.1 & 63.5 & 68.8 & 51.2 & 47.1 & 72.1 & 29.1 & 40.4 & 50.4 \\
    \midrule
    Qwen-GUI 
    \scriptsize w/ GUIAct 
    & 91.9 & 65.6 & 65.0 & \textbf{73.1} & \textbf{58.0} & \textbf{51.2} & 74.7 & \textbf{56.8} & \textbf{60.8} & 59.0 \\
    \small + GUIChat 
    & 90.9 & 69.4 & 66.7 & 68.9 & 52.5 & 46.8 & 73.0 & 55.7 & 58.1 & 57.2 \\
    \midrule
    MiniCPM-GUI 
    \scriptsize w/ GUIAct  & 89.1 & 52.4 & 55.2 & 62.3 & 42.0 & 48.6 & 64.5 & 29.6 & 43.3 & 49.0 \\
    \small + High Resolution  & 91.8 & 72.9 & 69.2 & 68.2 & 42.4 & 47.1 & 72.6 & 47.5 & 52.0 & 56.1 \\
    \small + GUIChat & 91.8 & 74.9 & 70.6 & 67.0 & 45.4 & 47.5 & 71.7 & 44.7 & 53.3 & 57.1 \\
    \small {\todot{+ Larger LM}} & \textbf{92.1} & \textbf{81.6} & \textbf{77.0} & {68.2} & 52.5 & 49.1 & \textbf{76.1} & 49.9 & 52.8 & \textbf{59.6} \\
    \bottomrule
  \end{tabular}}
  \caption{The performance of our GUI agents on test datasets. }
  \label{tab: guiact-additional-results}
\end{table*}

\section{Data Examples}
\label{sec: data-examples}

\begin{figure*}[tb]
  \centering
  \includegraphics[width=1.0\linewidth]{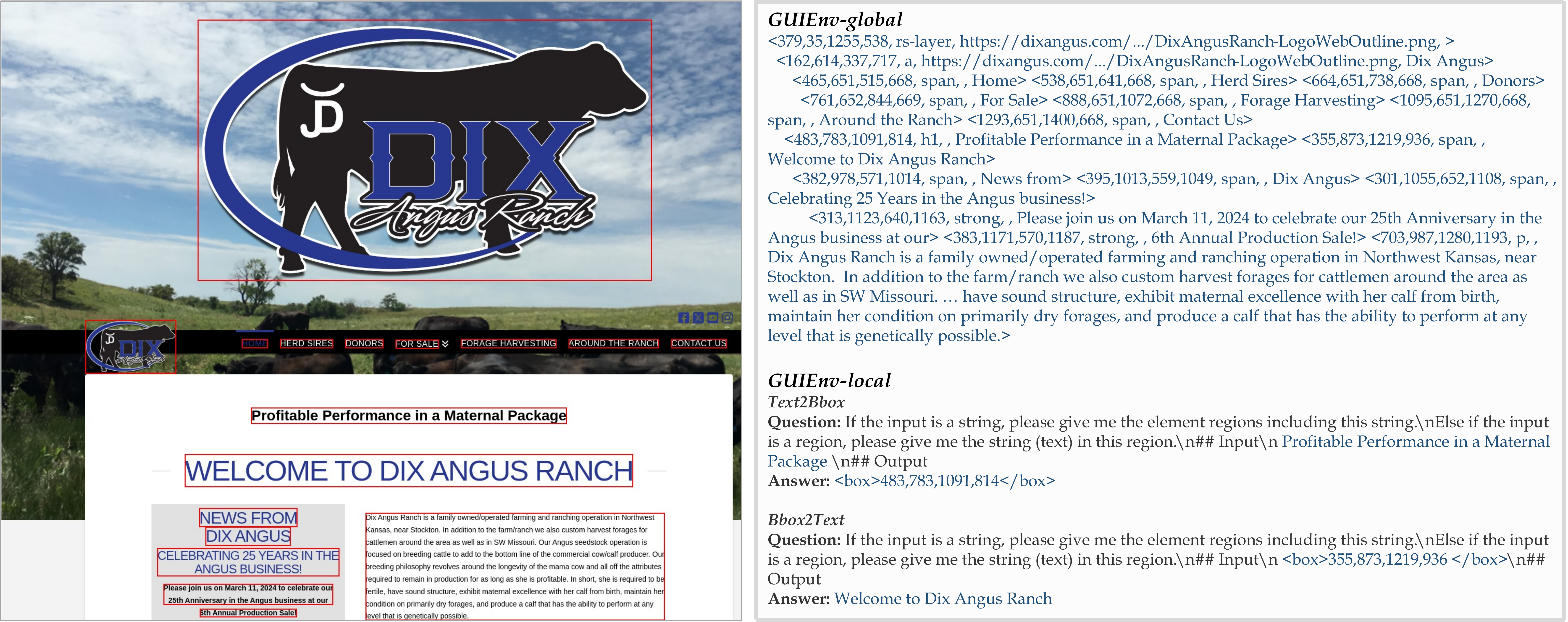}
  \caption{Some examples to show the data formats of our GUIEnv datasets. The GUIEnv-global task only provides a screenshot website, and the models need to predict all the texts and their positions according to the layout sequence. The GUIEnv-local task requires models to predict the region (a box represented by a left-top point and a right-down point) given the designated text or predict the text given the designated box.}
  \label{fig: GUIEnv-example}
\end{figure*}

\begin{figure*}[tb]
  \centering
  \includegraphics[width=0.95\linewidth]{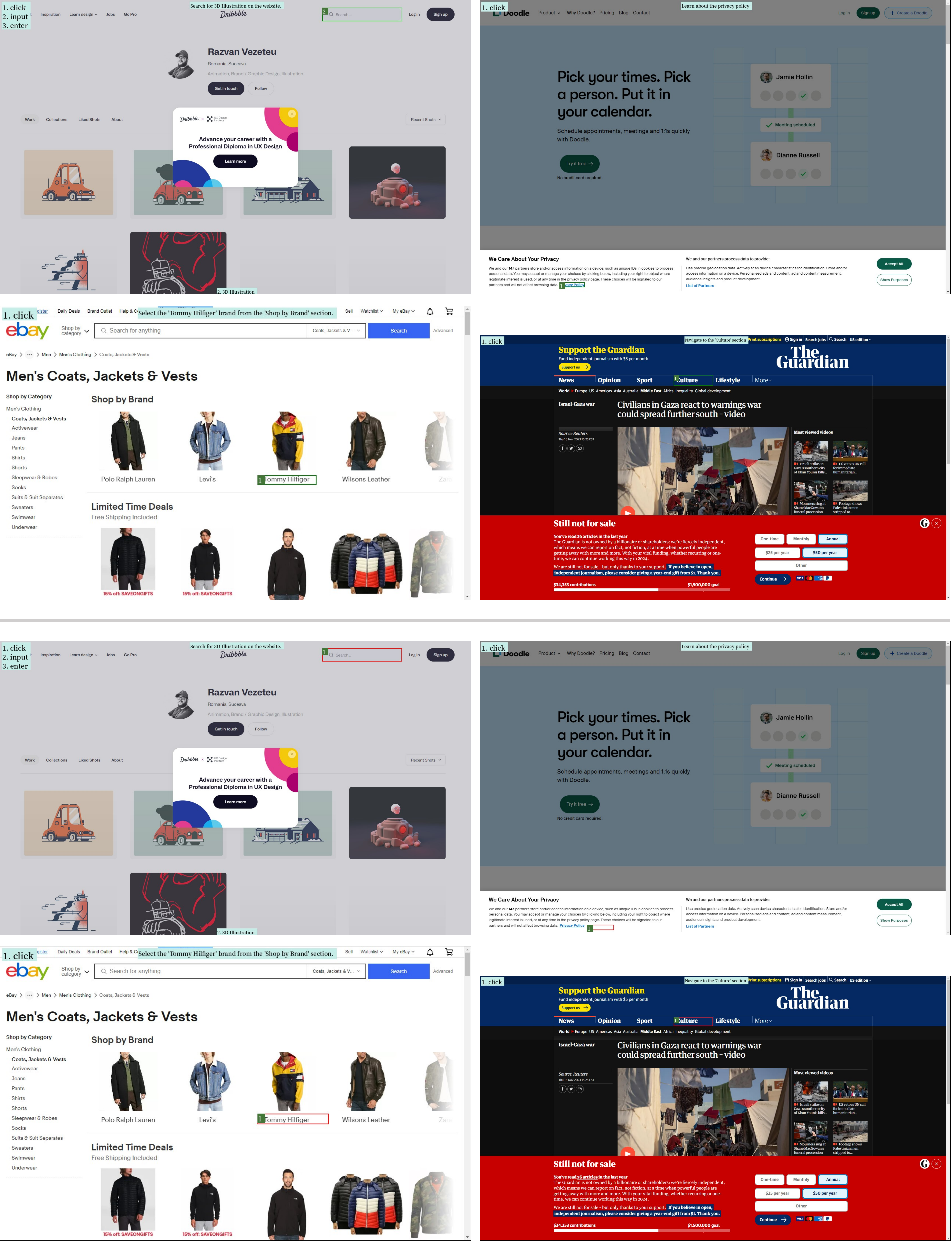}
  \caption{Four examples in the GUIAct-web-single datasets. The instructions are on the top of each screenshot, while the actions' names are on the left-top position. Moreover, we draw a box with an index to represent the "click" action. We use \textcolor{darkgreen}{green} color to visualize the \textcolor{darkgreen}{golden actions} and use \textcolor{red}{red} color to visualize the \textcolor{red}{predicted actions}.}
  \label{fig: GUIAct-web-single}
\end{figure*}

\begin{figure*}[tb]
  \centering
  \includegraphics[width=1.0\linewidth]{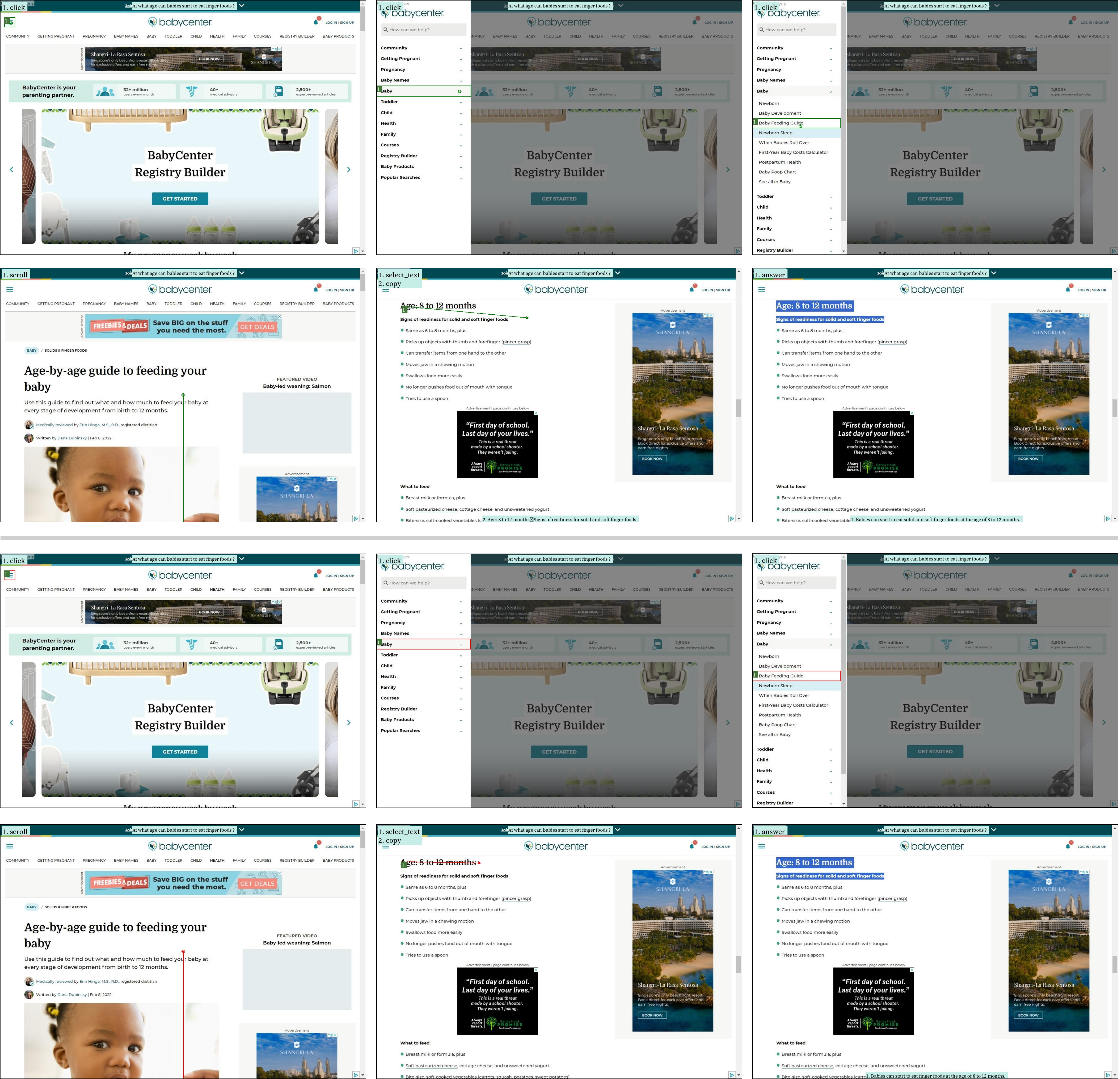}
  \caption{A 6-step navigation example of GUIAct-web-multi datasets. The instructions are on the top of each screenshot, while the actions' names in each step are shown in the left-top position. These graphs show the "click", "scroll", "select\_text", "copy" and "answer" actions. We use \textcolor{darkgreen}{green} color to visualize the \textcolor{darkgreen}{golden actions} in the top-6 graphs and use \textcolor{red}{red} color to visualize the \textcolor{red}{predicted actions} in the bottom-6 graphs.}
  \label{fig: GUIAct-web-multi}
\end{figure*}

\begin{figure*}[ht]
  \centering
  \includegraphics[width=0.82\linewidth]{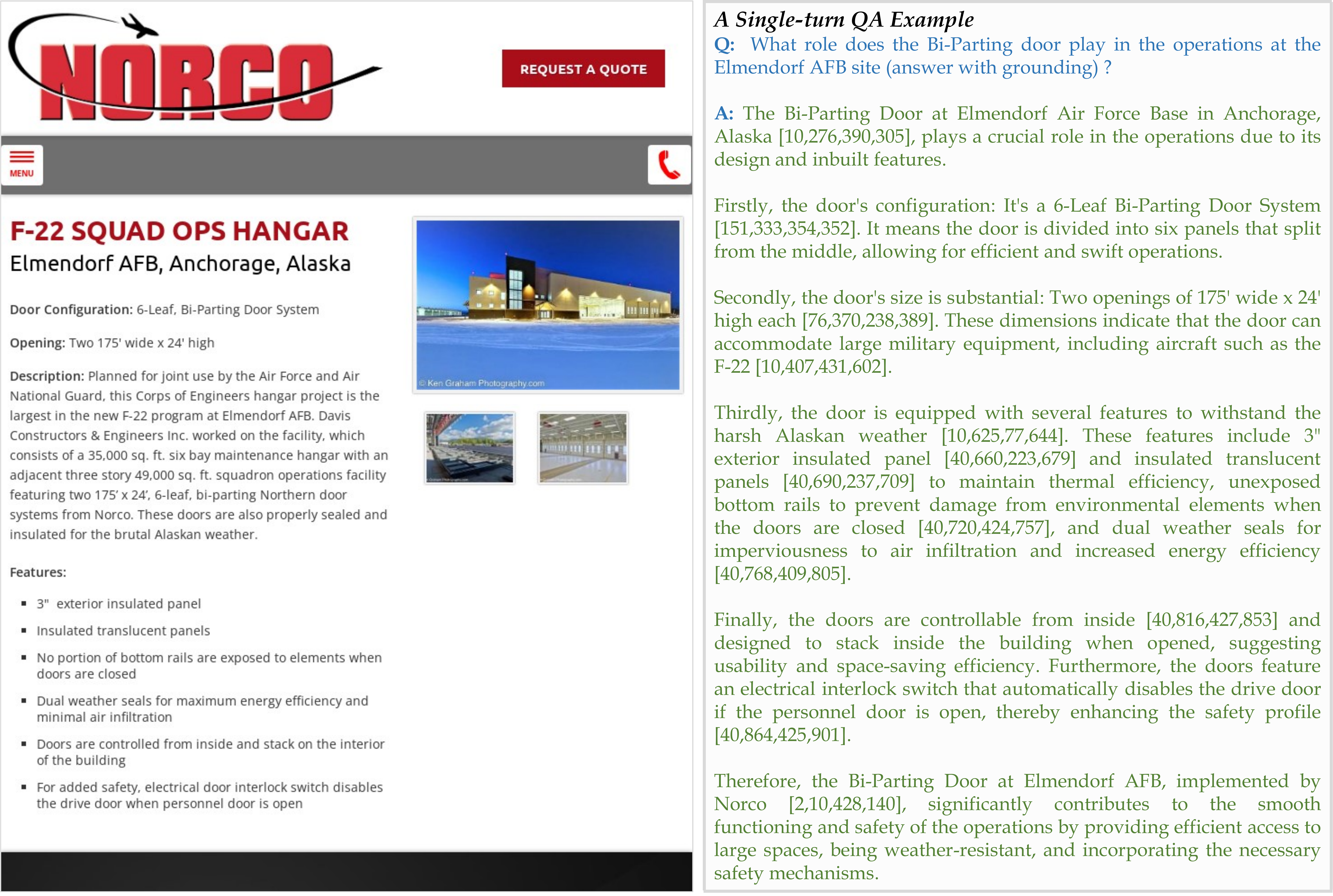}
  \caption{A case of GUIChat's single-turn QA.}
  \label{fig:guichat-singlecase}
\end{figure*}

\begin{figure*}[ht]
  \centering
  \includegraphics[width=0.82\linewidth]{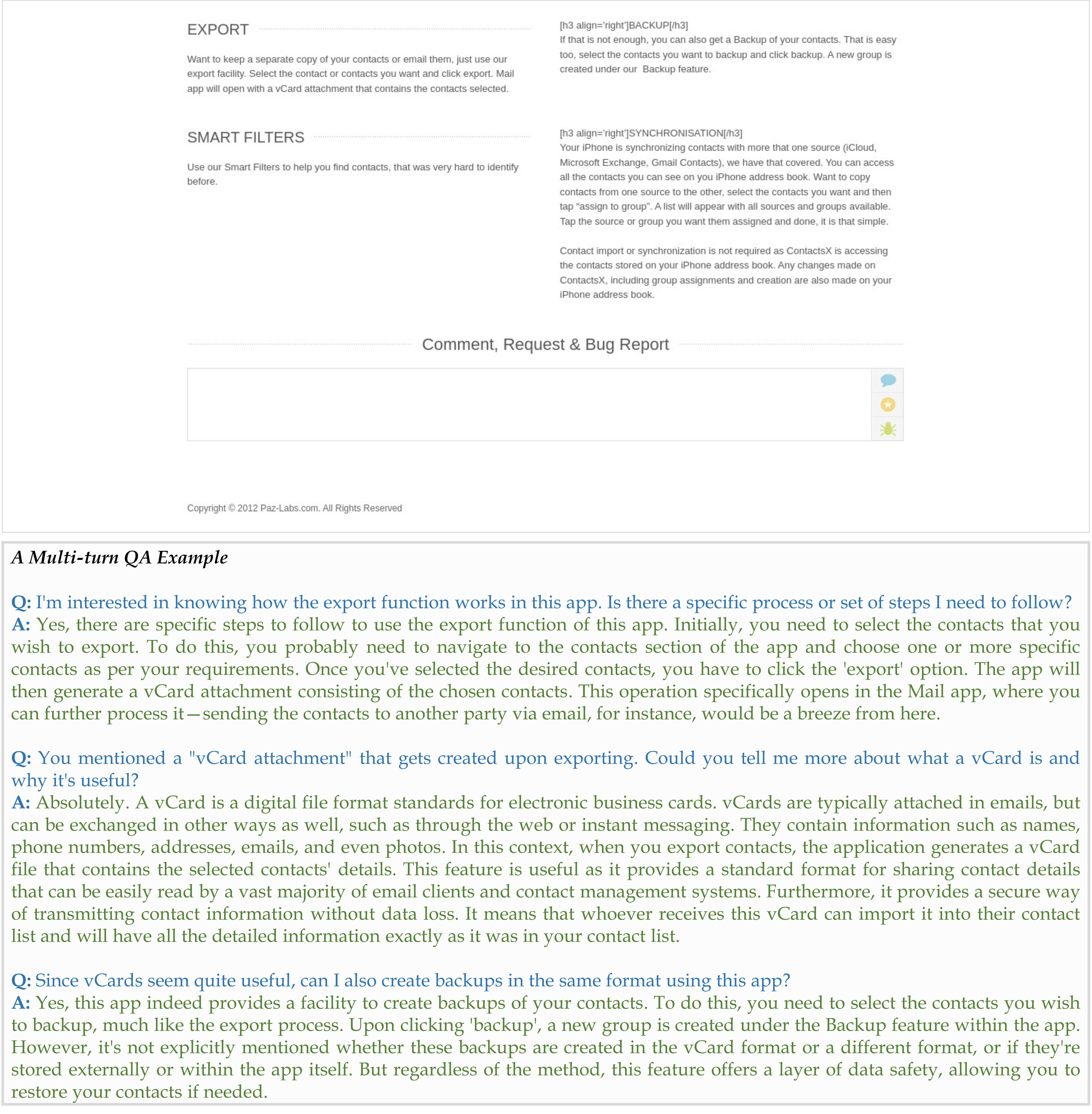}
  \caption{A case of GUIChat's multi-turn QA.}
  \label{fig:guichat-multicase}
\end{figure*}

We show some examples of GUICourse in this section. An example of GUIEnv-global and an example of GUIEnv-local are shown in Figure~\ref{fig: GUIEnv-example}. Some examples of GUIAct (web-single) are shown in Figure~\ref{fig: GUIAct-web-single}, and an example of GUIAct (web-multi) is shown in Figure~\ref{fig: GUIAct-web-multi}. A single-turn QA example of GUIChat is shown in Figure~\ref{fig:guichat-singlecase} and a multi-turn QA example of GUIChat is shown in Figure~\ref{fig:guichat-multicase}.

\section{Ethical Considerations and Societal Impacts}
\label{sec: limitation}

\paragraph{Ethical Considerations.} We provide fair wages for the annotators when constructing our datasets, and there is no personally identifiable information. When using existing assets, we cite the creators and use the open-sourced datasets or models.

\paragraph{Societal Impacts.} Our data in GUICourse don't need personal information or offensive content. Meanwhile, we don't select websites that might include personal information or offensive content. However, some screenshots in our GUIEnv dataset are collected from the Cleaned Common Crawl Corpus, so we cannot guarantee that these website screenshots are absolutely free of personally identifying information or offensive content.

\end{document}